\documentclass[a4paper,12pt]{article}
\usepackage{geometry}
\usepackage{multicol}
\usepackage{color}
\usepackage{xspace}
\usepackage{enumerate}
\usepackage{indentfirst}
\usepackage{algorithm}
\usepackage{algorithmic}
\usepackage{amsmath}
\usepackage{subfigure}
\usepackage{hyperref}
\usepackage{graphicx}
\usepackage{microtype}
\usepackage{caption}
\usepackage{multirow}
\usepackage{booktabs}
\usepackage{makecell}
\usepackage{times}
\usepackage{epsfig}
\usepackage{graphicx}
\usepackage{amsmath}
\usepackage{amssymb}
\usepackage{xcolor}
\usepackage{multirow}
\usepackage{makecell}
\usepackage{algorithm}
\usepackage{algorithmic}
\usepackage{bbm}
\usepackage{url}
\usepackage{adjustbox}
\usepackage{threeparttable}
\usepackage{setspace}
\usepackage[section]{placeins}
\date{}
\makeatletter

\begin{document}



\title{Explicitly Modeling the Discriminability for Instance-Aware Visual Object Tracking}
\author{Mengmeng~Wang,
		Xiaoqian~Yang,
		and~Yong~Liu
	}
%
\maketitle
\begin{abstract}
	 Visual object tracking performance has been dramatically improved in recent years, but some severe challenges remain open, like distractors and occlusions. We suspect the reason is that the feature representations of the tracking targets are only expressively learned but not fully discriminatively modeled. In this paper, we propose a novel Instance-Aware Tracker (IAT) to explicitly excavate the discriminability of feature representations, which improves the classical visual tracking pipeline with an instance-level classifier. First, we introduce a contrastive learning mechanism to formulate the classification task, ensuring that every training sample could be uniquely modeled and be highly distinguishable from plenty of other samples. Besides, we design an effective negative sample selection scheme to contain various intra and inter classes in the instance classification branch. Furthermore, we implement two variants of the proposed IAT, including a video-level one and an object-level one. They realize the concept of \textbf{instance} in different granularity as videos and target bounding boxes, respectively. The former enhances the ability to recognize the target from the background while the latter boosts the discriminative power for mitigating the target-distractor dilemma. Extensive experimental evaluations on 8 benchmark datasets show that both two versions of the proposed IAT achieve leading results against state-of-the-art methods while running at 30FPS. Code will be available when it is published.
	 
\end{abstract}

\section{Introduction}
Visual object tracking plays a vital role in many applications ranging from robotics to video surveillance. It aims at tracking target objects over video sequences with only an initial rectangle location given. There are several open challenges in visual tracking, including significant target variations, highly similar distractors and occlusions. Therefore, it is a challenging task to learn a robust representation that can handle these challenging scenarios. 

	\begin{figure*}[t]
		\begin{center}
			\includegraphics[width=0.9\linewidth]{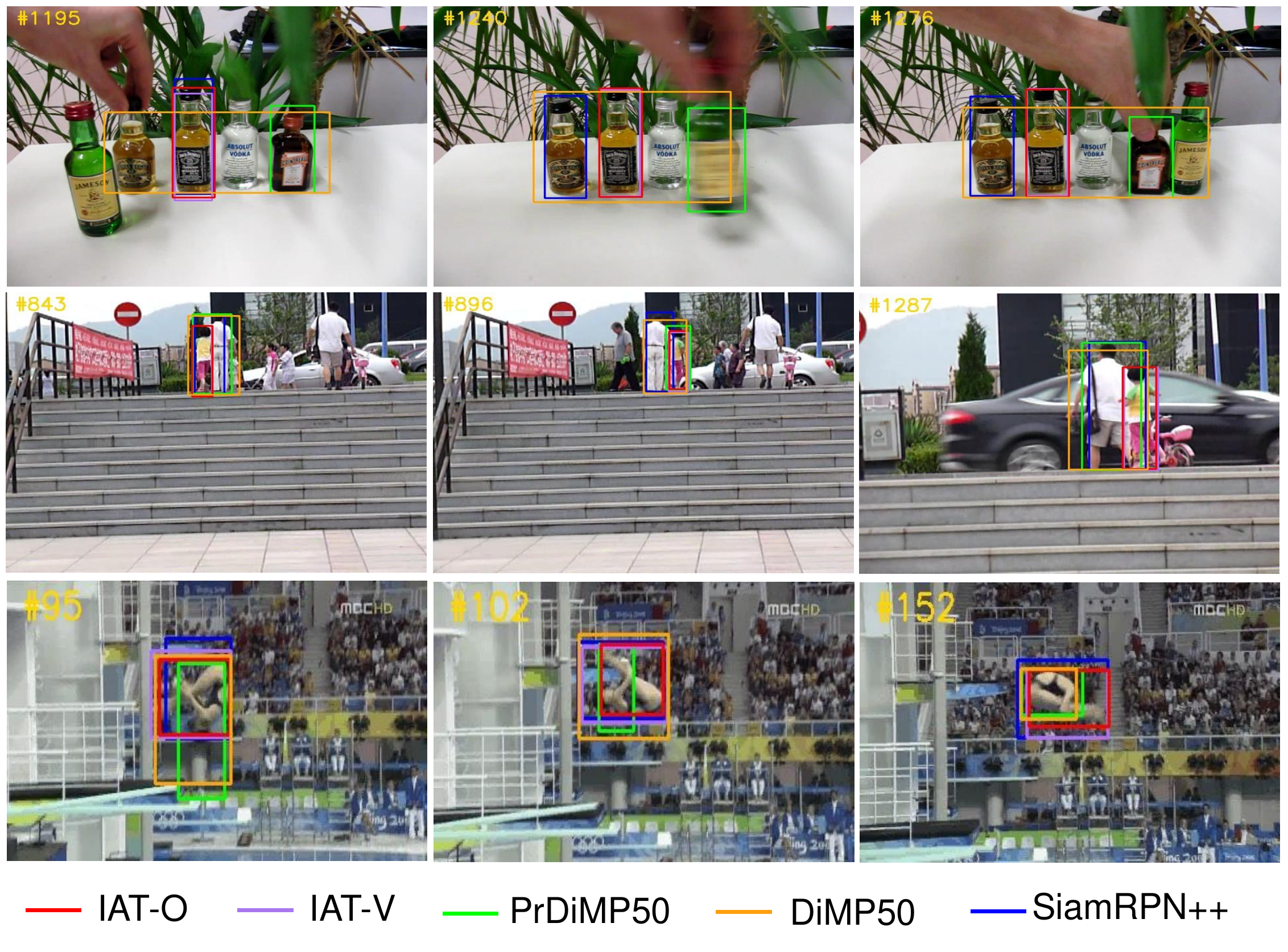}
		\end{center}

		\caption{Comparisons of the proposed video-level (IAT-V) and object-level (IAT-O) instance-aware trackers with the state-of-the-art trackers PrDiMP50~\cite{danelljan2020probabilistic}, DiMP50~\cite{bhat2019learning} and SiamRPN++~\cite{li2019siamrpn++}. Our trackers can accurately track targets even when objects suffer from distractors, severe occlusions and heavy deformation.}
		
		\label{fig:box}
	\end{figure*}
	
	Developing an expressive and discriminative  feature representation can dramatically improve the tracking performance~\cite{wangnaiyan}. It is proved by recent deep neural-network-based methods~\cite{danelljan2015convolutional,ma2015hierarchical}, which rely on the learned features from CNNs instead of the raw RGB images or the hand-crafted features. 
	Moreover, to further exploit the powerful expressive ability of CNNs, lots of end-to-end learning frameworks~\cite{danelljan2020probabilistic,cui2020fully,chen2020siamese,li2019siamrpn++} are proposed for visual tracking. Especially, siamese-network-based methods~\cite{li2018high,chen2020siamese,li2019siamrpn++}  have drawn great attention, which model tracking as a matching problem. They are devoted to designing a robust offline-training network for real-time visual tracking with different subnetwork branches like RPN~\cite{ren2015faster} to locate objects directly. The tracking results become more satisfactory since the end-to-end learning paradigms make the utmost use of the powerful representations of CNNs. However, even though they could handle lots of common problems in tracking, some severe challenges remain open like similar distractors and serious occlusions.
	
	Reminiscent of the way that humans track an object with our eyes, we actually capture information in two folds: (1) visual features that are directly perceived through our senses; (2) the distinctiveness of the tracking object from our cognition.
    For example, if we are required to track a white British cat, we will try our best to memorize several features like its eyes, color, ears' shape, etc. Moreover, one easily overlooked but the most important thing is that we have strong prior knowledge about this target. We know it is a specific cat and what is different between this cat and all the other cats, even all the other things in our mind. The distinction is the prerequisite of successful tracking. Our key insight is that a strong representation should depict the traits of a target not only \textit{expressively} but also \textit{discriminatively}. The former is achieved by employing deep and complicated networks, while the latter remains unsolved. There are indeed some approaches shedding some lights on this area. For instance, DaSiamRPN~\cite{zhu2018distractor} is proposed to improve the discriminative ability of SiamRPN~\cite{li2018high} by adding semantic negative sample pairs. However, this strategy has limited effect, since only one negative sample is seen at one time for one template input. SINT++~\cite{wang2018sint++} proposes to generate hard positive samples for robust tracking, but it has not paid enough effort on the importance of negative samples. Yao et al.~\cite{yao2021robust} selects hard negative samples to improve the model adaptation during online update, but using this strategy on inference phase will reduce the speed. DiMP~\cite{bhat2019learning} and PrDiMP~\cite{danelljan2020probabilistic} reach strong performance by fully exploiting the target-background difference, but their negative samples are limited. In a word, these methods model the discriminability either incompletely or implicitly, which hinders their further improvements. 
	
	Our idea is to explicitly model the discriminability for visual tracking by classifying every target as a specific category against plenty of the other samples. To achieve this goal, we introduce self-supervised contrastive learning~\cite{he2020momentum}, which is an excellent technique used to learn the general features without labels. It could teach the model to distinguish if the input samples are similar or different. Concretely, we propose an \textbf{I}nstance-\textbf{A}ware \textbf{T}racker (IAT), which reformulates the visual tracking problem as an instance-aware classification task. IAT explicitly maximizes its discriminative ability by designing an instance classification branch. This branch introduces a large memory bank to store plenty of both intra-class and inter-class negative samples and a memory encoder to extract the features of these samples. Then, the network is trained to contrastively distinguish the target object from lots of negative samples and correctly locate the object at the same time. Therefore the model could learn powerful representations for the object both discriminatively and expressively. Furthermore, there are two different levels of instance-aware features for visual tracking. The first is a video-level one, which regards a whole video as a sample. It could aggregate the ability to recognize the target from various backgrounds in a global view. The second one is object-level representations, which treats every target bounding box as an instance. It could enhance the discriminative power for mitigating the target-distractor dilemma. We implement the two instance granularities with two versions of IAT in this paper as IAT-V and IAT-O. Both of them have promising performance. We establish and integrate our instance-level classification task based on PrDiMP~\cite{danelljan2020probabilistic}. Note that the proposed instance classification branch is only used in the training process and disabled in the inference phase. Thus it would not increase any computational burden for online applications. As shown in Figure~\ref{fig:box}, both IAT-V and IAT-O achieve superior performance, which demonstrates the improved discriminative ability and powerful feature representations brought by our method. 
	
	In summary, our main contributions are as follows.
	\begin{itemize}
		\item A novel Instance-Aware Tracker (IAT) is proposed to explicitly model the discriminability of visual tracking, which incorporates a contrastive learning based instance-level classification task into the tracking pipeline. To the best of our knowledge, this is the first attempt to use contrastive learning in visual tracking.
		
		\item Two variants of instance feature representations including a video-level one and an object-level one are implemented in terms of different granularity to explore different discrimininative power of our trackers.
		
		\item We compare our method on 8 datasets with state-of-the-art methods. The experimental results and corresponding analysis demonstrate the effectiveness of the proposed method. It achieves leading results while running at 30FPS.
	\end{itemize}

\section{Related Works}
	We give a brief review about visual tracking from the perspective of feature representation. Readers can refer to a recent survey~\cite{kristan2020eighth} for detailed tracking development. Besides, we also introduce some relevant instance-aware trackers and self-supervised contrastive learning.

	\subsection{Feature Representation of Visual Tracking}
	

	Roughly speaking, visual tracking goes through three stages in recent years from the perspective of feature representation, including traditional handcrafted features, deep features and end-to-end deep features. 
	
	In the first stage, trackers usually use Haar~\cite{hare2015struck}, color~\cite{bertinetto2016staple} and HOG~\cite{wang2017large,henriques2014high,danelljan2015learning} to represent target objects and train a classifier to separate the targets from the backgrounds with many sophisticated techniques including multiple-instance learning~\cite{babenko2010robust}, structured output SVM~\cite{hare2015struck,wang2017large}, correlation filter~\cite{henriques2014high,danelljan2015learning,li2014scale,elayaperumal2021aberrance} and so on. The performance is limited by generalization capability of the features~\cite{wangnaiyan}. 
	
    Then the rapid development of deep neural networks pushes the visual tracking into the second stage. Several methods~\cite{danelljan2015convolutional,ma2015hierarchical} employ deep features into the existing tracking frameworks and dramatically improve the performance of the methods in the first stage. This phase lasts for a concise period since just substitute the traditional features with deep features extremely harms the tracking speed. 
	
	Now we are in the third stage, researchers try to take full advantage of deep neural networks by assembling feature extractors and classifiers into trainable end-to-end networks~\cite{nam2016learning,held2016learning,bertinetto2016fully,valmadre2017end,zhao2021deep}. Among them, siamese networks~\cite{he2018twofold,wang2018learning,li2018high,dong2018triplet,zhu2018distractor,zhang2019deeper} based methods have drawn great attention which model tracking as a template matching task and perform similarity learning. Recent prevalent topics that concentrated on this stage are various, like improving target-background discriminability~\cite{danelljan2019atom,bhat2019learning,zhao2021adaptive}, deeper and wider backbone for siamese networks~\cite{zhang2019deeper,li2019siamrpn++}, designing the siamese network with diverse components from object detections~\cite{guo2020siamcar,li2018high,xu2020siamfc++,chen2020siamese} and so on. The feature representation is further enhanced with these end-to-end networks which improve the expressive ability of deep features with elaborated classification and regression heads. Besides, many methods~\cite{ guo2021exploring, zheng2020multi} are dedicated to mining more effective and robust feature representations to improve tracking performance. On the other hand, there are more researchers focus on applying the spatial and temporal information simultaneously, and appear some works~\cite{zhang2020mining}, etc. These methods are all trying to use different feature extraction modules or fusion strategies to obtain the more effective and robust feature representations. In this work, we propose to dig the feature representation power deeper by exploring instance-aware features which have not been fully exploited in visual tracking.
	
	\subsection{Instance-aware Tracking}
	There are indeed several methods that model tracking as a special instance paradigm of some adjacent tasks like retrieval and detection. SINT~\cite{tao2016siamese} regards object tracking as an instance searching task by learning a robust matching function for matching arbitrary, generic objects. SINT++~\cite{wang2018sint++} is a follow-up of SINT, which introduces the positive samples generation network to sample massive diverse training data and the hard positive transformation network to generate hard samples. MAML~\cite{wang2020tracking} considers the tracking problem as a special type of object detection problem, i.e., instance detection. By employing meta-learning, the tracker can quickly learn how to detect a new instance using only one or few training samples from the initial or previous frames. Li et al.~\cite{li2019visual} investigate the problem of real-time accurate tracking in a instance-level tracking-by-verification mechanism and propose a multi-stream deep similarity learning network to learn a similarity comparison model purely offline. Quad~\cite{dong2019quadruplet} designs a shared network with four branches that receive a multi-tuple of instances as inputs and are connected by pair loss and triplet loss, according to the similarity metric, they select the most similar and the most dissimilar instances as the positive and negative inputs of triplet loss from each multi-tuple. There are also some works focusing on the instance-aware tracking in multi-person tracking task. Wu et al.~\cite{wu2019instance} learn instance-aware representations of the tracked person by a multi-branch neural network where each branch (instance-subnet) corresponds to an individual. Differently, our approach obtains instance-aware representation by directly classifying every target as a specific category against plenty of the other samples. Besides, we do not increase any burdens in the testing process but realize it all in the training phase.

	\subsection{Self-supervised Contrastive Learning}
	A comprehensive survey of self-supervised learning is beyond the scope of this paper, so we only briefly review discriminative self-supervised learning which is the most relevant to our work. Self-supervised learning focuses on training a model with no labeled data and looks into it for the supervisory signal to feed high capacity deep neural networks. Discriminative methods, especially contrastive methods~\cite{he2020momentum,grill2020bootstrap,chen2020simple} currently achieve the state-of-the-art performance in self-supervised learning. The contrastive approaches try to make the representation of different views of the same image close while spreading representations from different images apart~\cite{grill2020bootstrap}. What's in common with our motivation is that for contrastive methods and tracking, (1) using negative pairs could increase the discriminability of the model, (2) including more negative samples for every training sample could benefit more to the model. 

\section{Methodology}
In this section, we describe our proposed Instance-Aware Tracker (IAT) in detail. As shown in Figure~\ref{fig:framework}, the whole architecture of IAT consists of: feature extractors, a target classification branch, a bounding box regression branch and an instance classification branch. The three branches share two weight-shared feature extractors $f_{1}$ and $f_{2}$, where $f_{1}$ extracts the features for the template frames and $f_{2}$ works for the search frames. The classification branch predicts the center score map of a target in the search frame, considering the location with the maximum score as the center position of a bounding box. The regression branch estimates the bounding box of the target based on the predicted target center location. Moreover, the core instance classification branch classifies every target as a specific category against plenty of other samples to boost the discriminability of the tracker. 
	
    
    We will first briefly revisit the baseline tracker with the target classification and regression parts (Section~\ref{section3.1}); then we will detailedly introduce our core instance classification branch (Section~\ref{section3.2}) and present two variants of it (Section~\ref{section3.3});  finally we will introduce the whole training losses and process (Section~\ref{section3.4}).
	
	\subsection{Revisit of the Baseline Tracker} \label{section3.1}
    In this work, we consider a probabilistic regression tracker PrDiMP~\cite{danelljan2020probabilistic} as our baseline model. It models tracking with two subproblems, target classification and bounding box regression as most of the existing visual tracking networks~\cite{guo2020siamcar, chen2020siamese, bhat2019learning, danelljan2020probabilistic}. The two tasks as formulated as follows.\\
    
	\textbf{Target Classification:} Target classification branch aims to coarsely localize the target center in the image. Given a pair of features $(f_{1} \left (\textbf{I}_{t} \right ), f_{2} \left (\textbf{I}_{s} \right ))$, a target model $G$ will be employed to generate the filter weights $g=G(f_{1} \left (\textbf{I}_{t} \right ))$ of a correlation convolution layer. Then, the correlation weights $g$ is applied on the features of the search frame $f_{2} \left (\textbf{I}_{s} \right )$ to predict the target center location scores. The target classification loss is computed as the mean squared error of all search samples as:
	\begin{equation} \label{discriminative}
	\begin{aligned}
	L_{cls} = \frac{1}{N} \sum_{(\textbf{I}_{s},c) \in S} \parallel f_{2} \left (\textbf{I}_{s} \right ) * g - c \parallel ^2	+ \parallel \lambda g \parallel ^2
	\end{aligned}
	\end{equation}
	Here, $N$ is the length of the training set $S$, $*$ denotes the convolution operation and $\lambda$ is a regularization factor. $c$ is the ground truth target center coordinate.
	
   Besides, the target model $G$ can work for both offline training and online tracking. During online tracking process, given the first frame with annotation, the baseline constructs the initial target model training set $S$ by data augmentation~\cite{bhat2018unveiling}. The initial target model is then obtained using the steepest descent recursions. In the subsequent tracking procedure, the target model is easily updated by adding new training samples to the training set whenever a distractor peak is detected. The final target features are averaged over all samples in the training set, which is defined as the filter weights of the correlation convolution layer to discriminate between target and background appearance in the search frame feature space.\\
	
	\textbf{Bounding Box Regression:} This branch estimates the final bounding box location of the target based on the predicted target center location. There are various methods, usually including anchor-based IoU calculation~\cite{danelljan2019atom, bhat2019learning} and anchor-free width and height regression~\cite{chen2020siamese}. We use the same architecture as IoU-Net~\cite{danelljan2019atom} which is an anchor-based bounding box regression techniques, and apply the probabilistic regression formulation as \cite{danelljan2020probabilistic} to learn this branch, which proposes to probabilistically model label noise for the regression problem as a conditional ground-truth distribution $p(y| y_i)$, where $y$ and $y_i$ are the prediction and annotation, respectively. It generates a predictive probability distribution $p(y|x_i, \theta)$ of the output $y$ given the input $x_i=f_{2} \left (\textbf{I}_{s} \right )$ and a regression model with the parameters $\theta$. The network is trained to minimize the Kullback-Leibler divergence between $p(y|x_i, \theta)$ and $p(y| y_i)$: 
	\begin{equation} \label{probabilistic}
	\begin{aligned}
	L_{reg} = \int p(y | y_i)log\frac{p(y | y_i)}{p(y | x_i, \theta)}dy 
	\end{aligned}
	\end{equation}

	\subsection{Instance Classification Branch} \label{section3.2}
		\textbf{Instance Classification:} We propose to model the classical visual tracking problem as an instance-level classification task to explicitly model the discriminability of feature representations, achieving an instance-aware tracker. Our point is that a strong representation can be learned if (1) every training sample is uniquely modeled; (2) plenty of negative samples are involved for every training sample. We introduce self-supervised learning methods~\cite{he2020momentum,chen2020improved} to formulate this branch as a dictionary look-up task, regarding frames from the same video as positive samples while frames from other videos as negative samples.
	\begin{figure}[t]
		\begin{center}
			\includegraphics[width=1.0\linewidth]{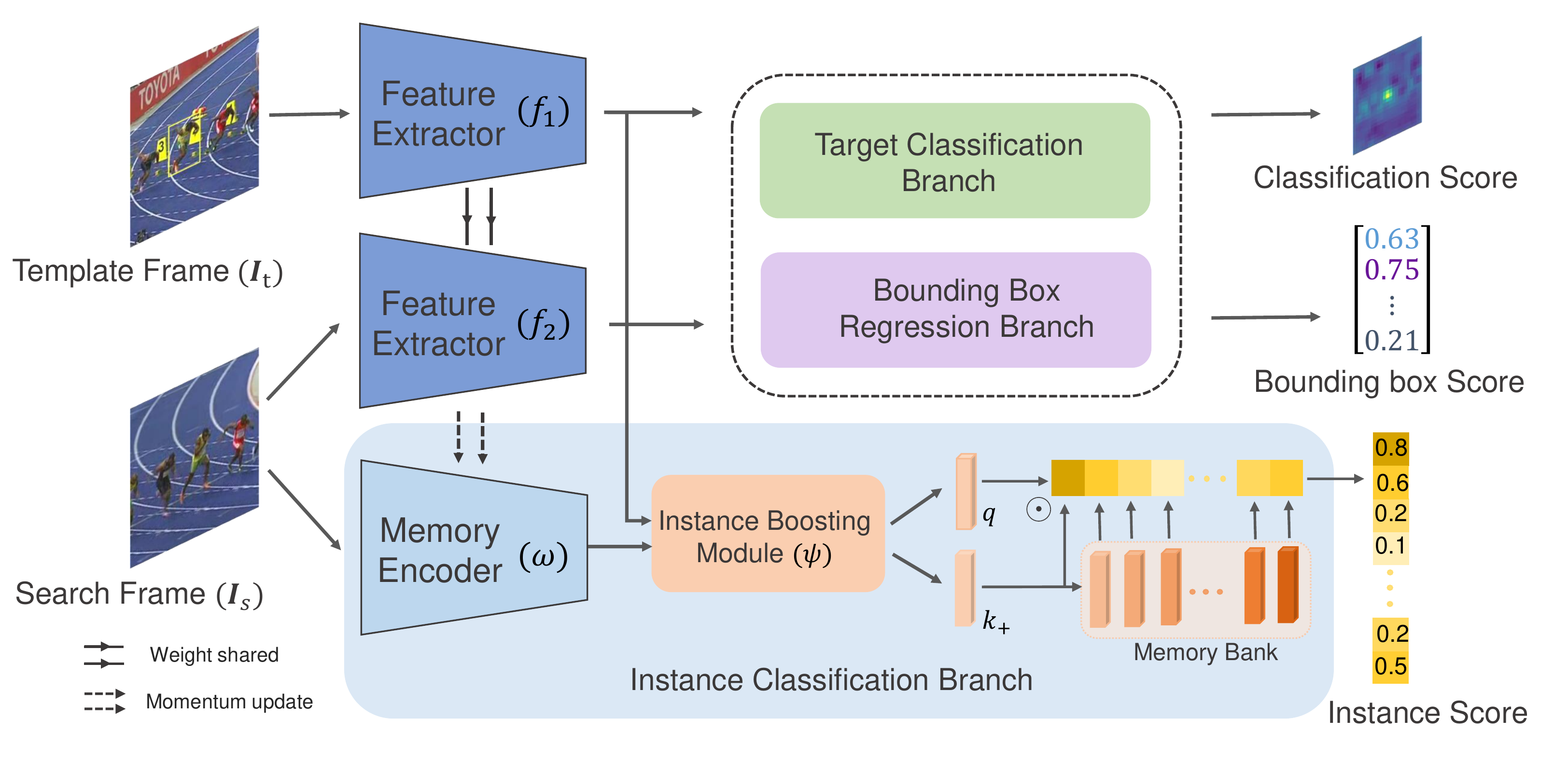}
		\end{center}
		\caption{\textbf{Instance-Aware Tracking Framework}. Given a template frame $\textbf{I}_{t}$ and a search frame $\textbf{I}_{s}$, we extract features using two shared-weights feature extractors $f_{1}$ and $f_{2}$ followed by three task-specific subnetworks, i.e., a target classification branch, a bounding box regression branch and an instance classification branch. The instance classification branch includes an instance boosting module $\psi$ and an extra memory encoder $\omega$ that is updated in a momentum way to encode the feature of $\textbf{I}_{s}$.}
		\label{fig:framework}
	\end{figure}
	
	As shown in Figure~\ref{fig:framework}, the instance classification branch includes a memory encoder $\omega$ and an instance boosting module $\psi$, where $\omega$ shares the same structure with $f_{1}/f_{2}$ and is used to extract the features of the dictionary, and $\psi$ is used to adjust features from $\omega$ and $f_{1}$ for contrastive learning. Given a query $q$ and a dictionary $\left \{ k_{+}, \{\mathrm{k}_{i}\}_{i=1}^{K} \right \}$ which contains a single positive key $k_{+}$ and a large amount ($K$) of negative keys, the contrastive learning task is to correctly look up $k_{+}$ for $q$ from the dictionary. In our case, $q$ corresponds to the feature of the input template frame, $k_{+}$ is the feature of the input search frame and $\{\mathrm{k}_{i}\}_{i=1}^{K}$ is the other historical search frames from other videos. During the training process, given the template frame $\mathbf{I}_{t}$ and the search frame $\mathbf{I}_{s}$ from the same video, the proposed instance classification branch goes forward:
	\begin{equation} \label{instance}
	\begin{aligned}
	q=\psi\left ( f_{1} \left (\textbf{I}_{t} \right )\right ) ,\\
	k_{+}=\psi\left ( \omega \left ( \textbf{I}_{s} \right )\right )
	\end{aligned}
	\end{equation}
	
	The dictionary is a large memory bank and implemented as a queue to reuse the encoded features from historical search frames. In other words, there is no need to recalculate the features $ \{\mathrm{k}_{i}\}_{i=1}^{K}$ at every iteration but enqueue current $k_{+}$ into the memory bank and dequeue the oldest feature after the iteration. Then the dynamic memory bank can involve a lot of keys and cover a rich set of negative samples. To make the comparisons of the keys to the query consistent, the encoder $\omega$ for the dictionary keys should be kept as consistent as possible with the encoder $f_{1}$ for the query. Therefore, $\omega$ is momentum updated from $f_{1}/f_{2}$ (weight-shared):
	\begin{equation}\label{momentum}
	\omega\leftarrow \lambda \omega + \left ( 1- \lambda\right )f_{1},\omega_{0}=f_{1}^{0}
	\end{equation}
	where $\lambda$ is a momentum coefficient, the initial parameters of $\omega_{0}$ are given by $f_{1}^{0}$ which is the initial weights of $f_{1}$. The instance boosting module $\psi$ is updated by back-propagation and its structure will be explained in Section~\ref{section3.3}. 
	
	The similarity of samples is measured by dot product and the final contrastive loss function is like the InfoNCE~\cite{oord2018representation} as follows:
	\begin{equation} \label{contrastive}
	L_{ins} = -\mathrm{log}\frac{\mathrm{exp}\left ( q\cdot k_{+} /\tau \right )}
	{\sum _{i=0}^{K}\mathrm{exp}\left ( q\cdot k_{i} /\tau   \right )}
	\end{equation}
	where $\tau $ is a temperature hyper parameter and $ \left ( K+1  \right )$ is the size of memory bank with $K$ negative samples and one current positive sample $k_{+}$ enqueued in the end. Intuitively, this loss is a log loss of a $ \left ( K+1  \right )$-way softmax-based classifier that tries to classify $q$ as $k_{+}$. \\
	\begin{figure*}[h]
		\begin{center}
			\includegraphics[width=0.7\linewidth]{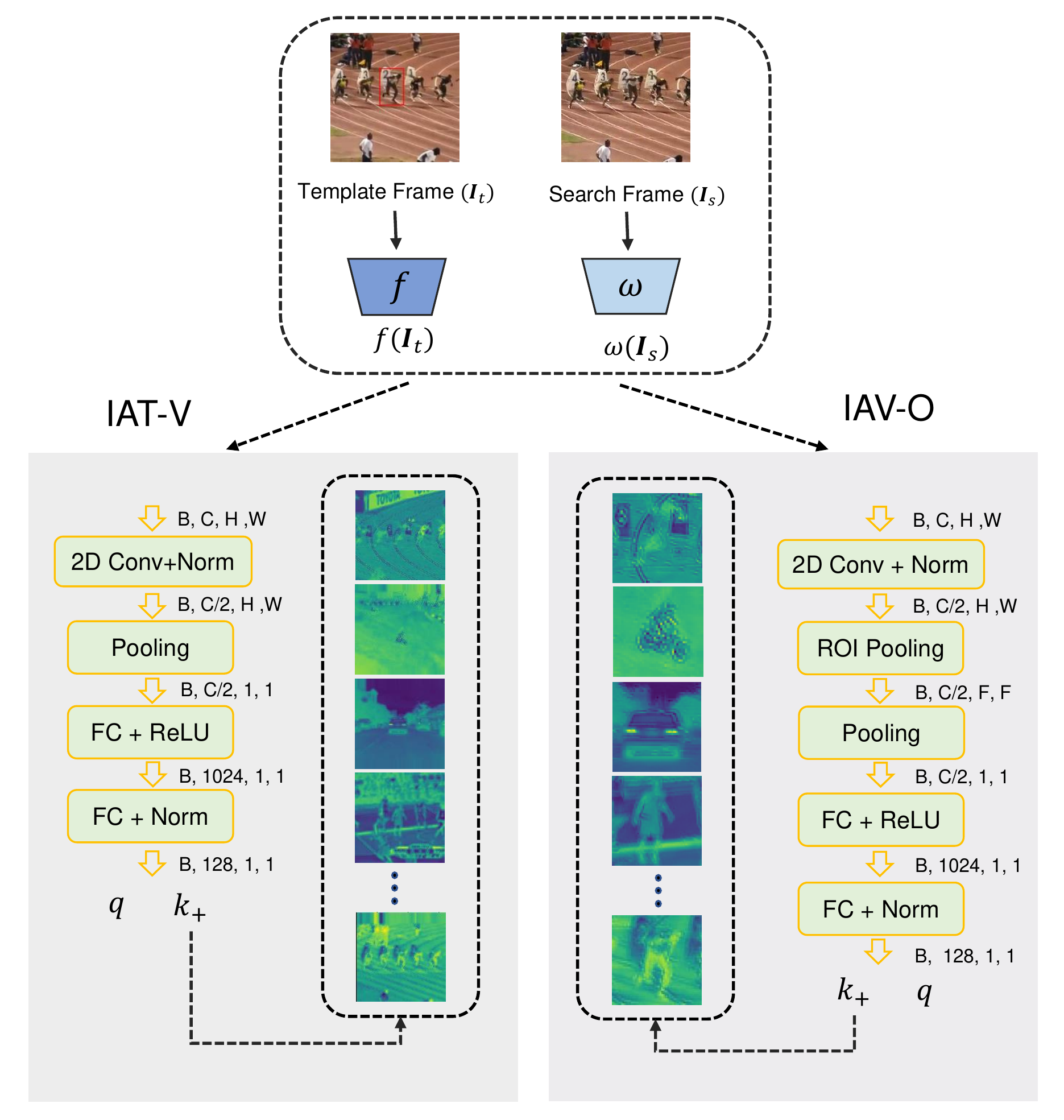}
		\end{center}
		\caption{\textbf{Differences of IAT-V and IAT-O}. The structures of the instance boosting module (IBM) $\psi$ and the memory bank of IAT-V (left) and IAT-O (right) are presented. Note that the features are extracted from a low-level layer for better visualization. Best view in color.}
		\label{fig:levels}
	\end{figure*}
	
	\textbf{Negative Selection Scheme:} The selection strategy of negative samples in contrastive learning is an important step. In our scheme, we treat the targets from the same sequence as the positive samples, but all the other samples from different sequences as negative samples. All the negative samples are maintained with a large memory bank. When the memory bank size is large enough, it could contain negative samples from both the same semantic category and different categories. Rather than only selecting negative samples of the same category, diversity of negative samples enables the instance classification branch to enhance both the inter-class and intra-class discriminability. 
	
	It is worth to be noted that the proposed instance classification branch is only enabled in the training process to explicitly improve the discriminative ability of IAT. It will be silent in the inference phase, thus it would not increase any computational burden for online applications.

	\subsection{Two Variants of IAT} \label{section3.3}
	We implement two variants of our method by realizing the concept of \textbf{instance} in different granularity, including a video-level one (IAT-V) and an object-level one (IAT-O). The differences between them are presented in the instance boosting module (IBM) $\psi$ and the obtained features.


	IAT-V treats every video as an instance. It uses the features of the entire input frame which contain both the target and background. We showcase the features and the structure in $\psi$ of IAT-V in the left part of Figure~\ref{fig:levels}. Specifically, $\psi$ consists of a 3x3 convolutional layer, a global pooling layer and two fully connected layers in this variant. We reduce the spatial channels by a factor of 2 to ease the computing cost in the 3x3 convolutional layer. The features become 128-dimensional vectors after the fully connected layers and we put $k_{+}$ into the memory bank. The dissimilarity between videos can be obtained from the entire frame features in the video-level granularity. The target in a certain video will have a chance to see and learn the target-distractor differentiation against both the targets and backgrounds in other negative videos. Therefore, IAT-V could enhance the ability to recognize the target from various backgrounds which may include both semantic objects or non-semantic clutter in a global view. 

	IAT-O regards every target bounding box as an instance. It is formed from a more concrete perspective that distincts characteristics existing between different objects. As shown in the right part of Figure~\ref{fig:levels}, we add a ROI Pooling layer after the 3x3 convolutional layer to obtain the object feature $\textbf{H} \in \mathcal{R}^{B, C/2, F, F}$, where the $F$ is the filter size of ROI Pooling layer. Then the features $\textbf{H}$ are fed into the latter layers and finally get 128-dimensional vectors. Then the vector $k_{+}$ is enqueued into the memory bank. Different from IAT-V, the vast negative samples here are all semantic object instances. Thus IAT-O could aggregate the discriminative power for easing the target-distractor dilemma. 
	
	
	
	
	\subsection{Training Loss} \label{section3.4}
	The loss function in our method consists of three terms corresponding to the three branches: 
	\begin{equation} \label{losses}
	\mathcal{L}=\lambda_{cls} L_{cls}+\lambda_{reg} L_{reg}+\lambda_{ins} L_{ins}
	\end{equation}
	where $\lambda_{cls}$, $\lambda_{reg}$ and $\lambda_{ins}$ are the coefficients used to balance the three loss terms. $L_{cls}$ is the classification loss in Eq.~(\ref{discriminative}) to supervise the target classification branch, $L_{reg}$ is the regression loss in Eq.~(\ref{probabilistic}) to supervise the regression branch, and $L_{ins}$ is from Eq.~(\ref{contrastive}) to supervise the instance classification branch. 
	With the above description of our instance-aware tracker (IAT), we present the whole training process in Algorithm~\ref{alg:algorithm}.
	
	\begin{algorithm*}
	\centering
		\caption{Pseudo code of the training process of IAT.}
		\label{alg:algorithm}
		\begin{algorithmic}[1]
			\STATE {Initialize the two shared feature extractors $f_1$, $f_2$ and memory encoder $\omega$, where $ \omega_{0} = f_{1}^{0}$.}
			\STATE {Initialize the features in the memory bank randomly.}
			\FOR{$\textbf{I}_{t}$ and $\textbf{I}_{s}$ in loader}
			\STATE forward $cls(f_{1}\left ( \textbf{I}_{t}  \right ), f_{2} \left ( \textbf{I}_{s}  \right ))$ and obtain $L_{cls}$ as Eq.\ref{discriminative}.
			\STATE forward $reg( f_{1}\left ( \textbf{I}_{t}  \right ), f_{2}\left ( \textbf{I}_{s}  \right ))$ and obtain $L_{reg}$ as Eq.\ref{probabilistic}.
			\STATE obtain q = $\psi(f_{1} (\textbf{I}_{t}))$.
			\STATE obtain $k_{+}$ = $\psi (\omega (\textbf{I}_{s}))$.
			\STATE $queue$: extract from memory bank.
			\STATE dequeue the oldest one in the $queue$ and enqueue $k_{+}$.
			\STATE calculate $L_{ins}$ as Eq.\ref{contrastive}.
			\STATE calculate $\mathcal{L}$ as Eq.\ref{losses}.
			\STATE update all parameters except $ \omega $ by Adam.
			\STATE update parameters of $\omega$ from $f_{1}$ or $f_{2}$ (weight-shared): $\omega\leftarrow \lambda \omega + \left ( 1- \lambda\right )f_{1}$.
			\ENDFOR
		\end{algorithmic}
	\end{algorithm*}

\section{Experiments and Discussion}
In this section, we first give the implementation details of our method. Then, we conduct ablation studies to analyse the influence and settings of instance classifier. Then, we compare our method with state-of-the-art trackers on 8 benchmarks including OTB50~\cite{wu2013online}, OTB100~\cite{wu2013online}, GOT-10k~\cite{huang2019got}, LaSOT~\cite{fan2019lasot}, Need for Speed (NFS)~\cite{kiani2017need}, UAV123~\cite{mueller2016benchmark}, TrackingNet~\cite{muller2018trackingnet} and VOT2019~\cite{kristan2019seventh}.

\subsection{Implementation Details}
	Our tracker IAT is implemented in Python using PyTorch. The training datasets we use include the training splits of GOT-10k~\cite{huang2019got}, LaSOT~\cite{fan2019lasot}, TrackingNet~\cite{muller2018trackingnet} and COCO~\cite{lin2014microsoft}. 40,000 videos are sampled per epoch from 4 datasets. The input template frames and the search frames are extracted with a random translation and scale relative to the target annotation. The backbones of the shared feature extractors $f_{1}$ and $f_{2}$ are ResNet50 pre-trained on ImageNet. We train our whole model for 50 epochs using Adam with a learning rate decay of 0.2 at the epoch of 15, 30 and 45, spending around 48 hours on 4 GeForce GTX 1080ti GPUs. The coefficients of the three branches are $\lambda_{ins}=0.01$, $\lambda_{cls}=100$ and $\lambda_{reg}=0.01$. We set $K$=1000 for the size of memory bank and $\lambda=0.999$ to update the memory encoder $\omega$. For IAT-O, we set the filter size $F$=3. In the inference phase, the average tracking speed is over 30 FPS on a single 1080ti GPU. 
	
		\begin{table}
			\begin{center}
	
				\scalebox{0.7}{
	
					\begin{tabular}{c| c| c| c| c| c| c| c}
						\hline
						\multicolumn{5}{c|}{$K$} &  \multirow{2}*{OTB50(\%)}  &  \multirow{2}*{OTB100(\%)} & \multirow{2}*{Combined(\%)}\\
						\cline{1-5}
						0 &10 &100 &1000 &10000            & & &\\	\hline
						\checkmark& &  &  &            & 67.73 & 70.16 & 67.2 \\
						&\checkmark &  &  &           & 68.21 & 71.07 & 67.07\\
	                    & &\checkmark  &  &           & 69.43  & 71.35 & 67.77 \\
						& &   &\checkmark &           & \textbf{70.28}  & \textbf{71.62} & \textbf{68.5} \\
						& &   & &\checkmark           & 69.35  & 71.11 & 67.64\\
						\hline
					\end{tabular} 
				}
			\end{center}
			\caption{Ablation study of the memory bank size $K$ in IAT-O. Combined means the large OTB100-NFS30-UAV123 dataset.}
			\label{tab:ablation-icb}
		\end{table}
		\begin{table}
			\begin{center}
	
				\scalebox{0.8}{
	
					\begin{tabular}{c| c| c| c| c| c| c| c}
						\hline
						\multicolumn{5}{c|}{$F$} &  \multirow{2}*{OTB50(\%)}  &  \multirow{2}*{OTB100(\%)} & \multirow{2}*{Combined(\%)}\\
						\cline{1-5}
						3 &5 &7 &9 &11            & & & \\	\hline
						\checkmark& &  &  &            & 70.28 & \textbf{71.62} & \textbf{68.5} \\
						&\checkmark &  &  &           & \textbf{70.29} & 71.36 & 67.77\\
	                    & &\checkmark  &  &           & 69.73  & 71.05 & 67.55 \\
						& &   &\checkmark &           & 68.97  & 70.85 & 67.42\\
						& &   & &\checkmark           & 68.47  & 70.44 & 67.10\\
						\hline
					\end{tabular} 
				}
			\end{center}
			\caption{Ablation study of the filter size $F$ of the ROI Pooling layer in IAT-O. Combined means the large OTB100-NFS30-UAV123 dataset.}
			\label{tab:ablation-roi}
		\end{table}
	\begin{table}
		\begin{center}
			\scalebox{1}{
				\begin{tabular}{c|c| c| c }
					\hline 
					Fusion Method & OTB50(\%) & OTB100(\%) & Combined(\%) \\
					\hline 
					IAT-V & 69.21 & 71.19 & 67.9 \\
					\hline
					IAT-O & 70.28 & 71.62 & 68.5 \\
					\hline
					Fusion(shared) & 70.27 & 71.61 & 68.4 \\
					\hline
					Fusion(separated) & 70.30 & 71.63 & 68.5 \\
					\hline
			\end{tabular} }
		\end{center}
		\caption{Ablation study of the different fusion method of IAT-V and IAT-O. Combined means the large OTB100-NFS30-UAV123 dataset.}
		\label{tab:ablation-fusion}
	\end{table}
	
	\subsection{Ablation Study}\label{absection}
	We perform the ablation study to demonstrate our proposed method on OTB50, OTB100 as well as a large combined OTB100-NFS30-UAV123 dataset. Unless specified, we use IAT-O in this section. The AUC of success is reported.\\
	
	\textbf{Memory Bank Size:}
	To explore the influence of the memory bank size $K$ in the proposed instance classification branch, we conduct experiments as shown in Table~\ref{tab:ablation-icb}. The first row with $K$=0 is the results of our basic pipeline. We can prove our point that using plenty of negative samples for every training sample is beneficial on all three datasets. When we increase $K$, the performance begins to be improved but obtains its peak at $K=1000$. We think the reason is that the oversize memory bank may make it too hard to classify well, making the performance drops. Thus we finally set $K$ to 1000 to ensure the best performance.\\
	
	\textbf{Filter Size:}
	We investigate the impact of filter size $F$ used in the ROI Pooling layer as suggested in Table~\ref{tab:ablation-roi}. It shows that $F$ has obvious impacts on the results. The results of $F$=3 and $F$=5 are almost the same on OTB50 dataset, but the gap is evident on the large combined dataset, thus we consider $F$=3 gives the best performance. \\
	
	\textbf{Combination of IAT-V and IAT-O:}
	IAT-V and IAT-O are mainly different in the concept granularity of instance, where the former treats every video as an instance while the latter regards every target bounding box as an instance. We try two fusion strategies to explore the combined performance IAT-V and IAT-O as shown in Table~\ref{tab:ablation-fusion}. $Fusion(shared)$ means the two variants share the same instance classification branch and their embedded features are added together for contrastive learning. $Fusion(separated)$ means the instance classification branch are not shared and the embedded features are also added. The results demonstrate that the first fusion strategy brings a slight performance drop compared to IAT-O, and the second strategy improves a little over IAT-O while brings extra parameters for training. Therefore, we think it is unnecessary to combine the two variants together and conduct all the comparative experiments with the independent version.

	\begin{figure}[t]
		\centering
		\begin{minipage}[b]{0.5\linewidth}
			\centering
			\centerline{
				\includegraphics[width=1.0\linewidth]{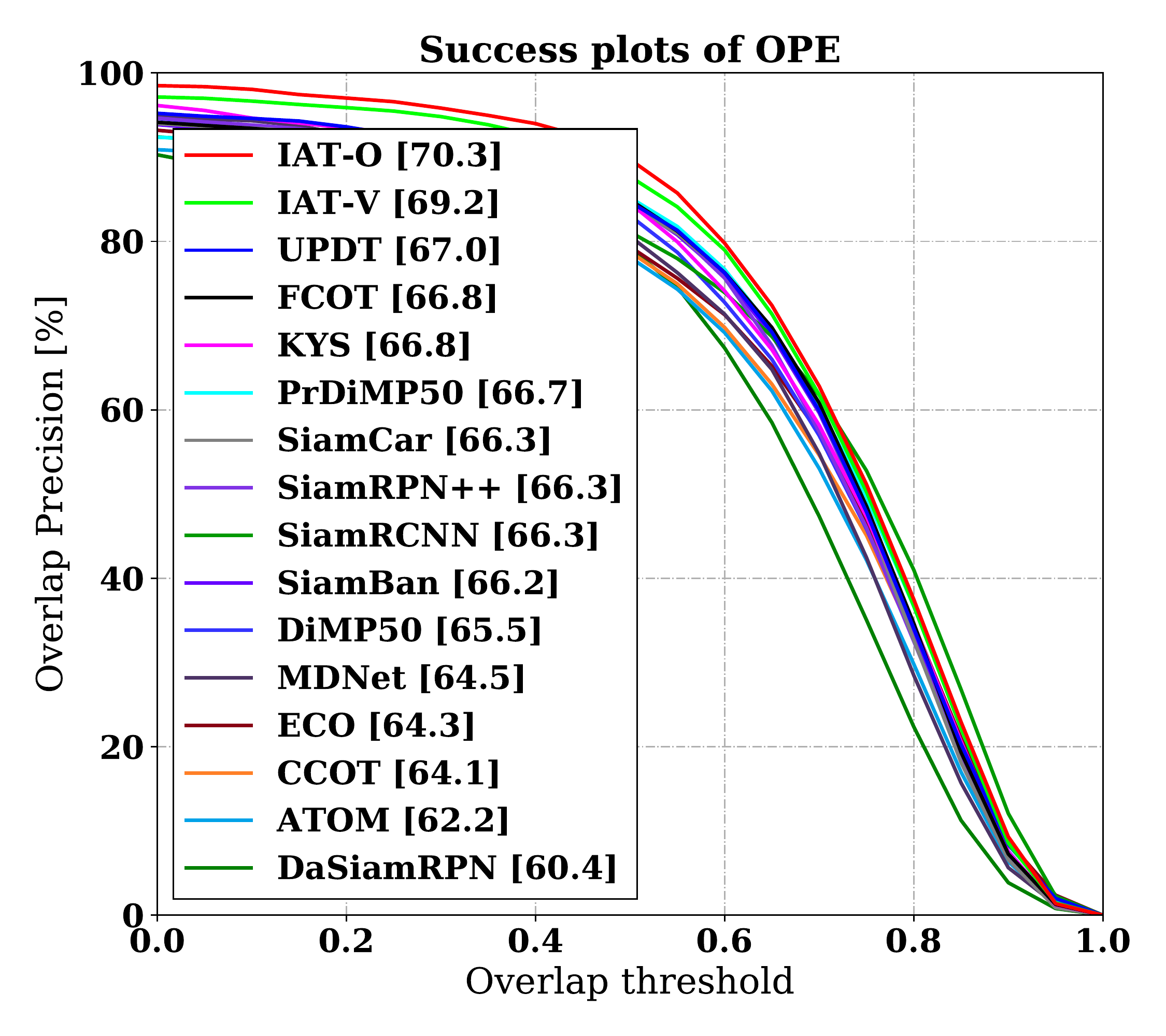}
				\includegraphics[width=1.0\linewidth]{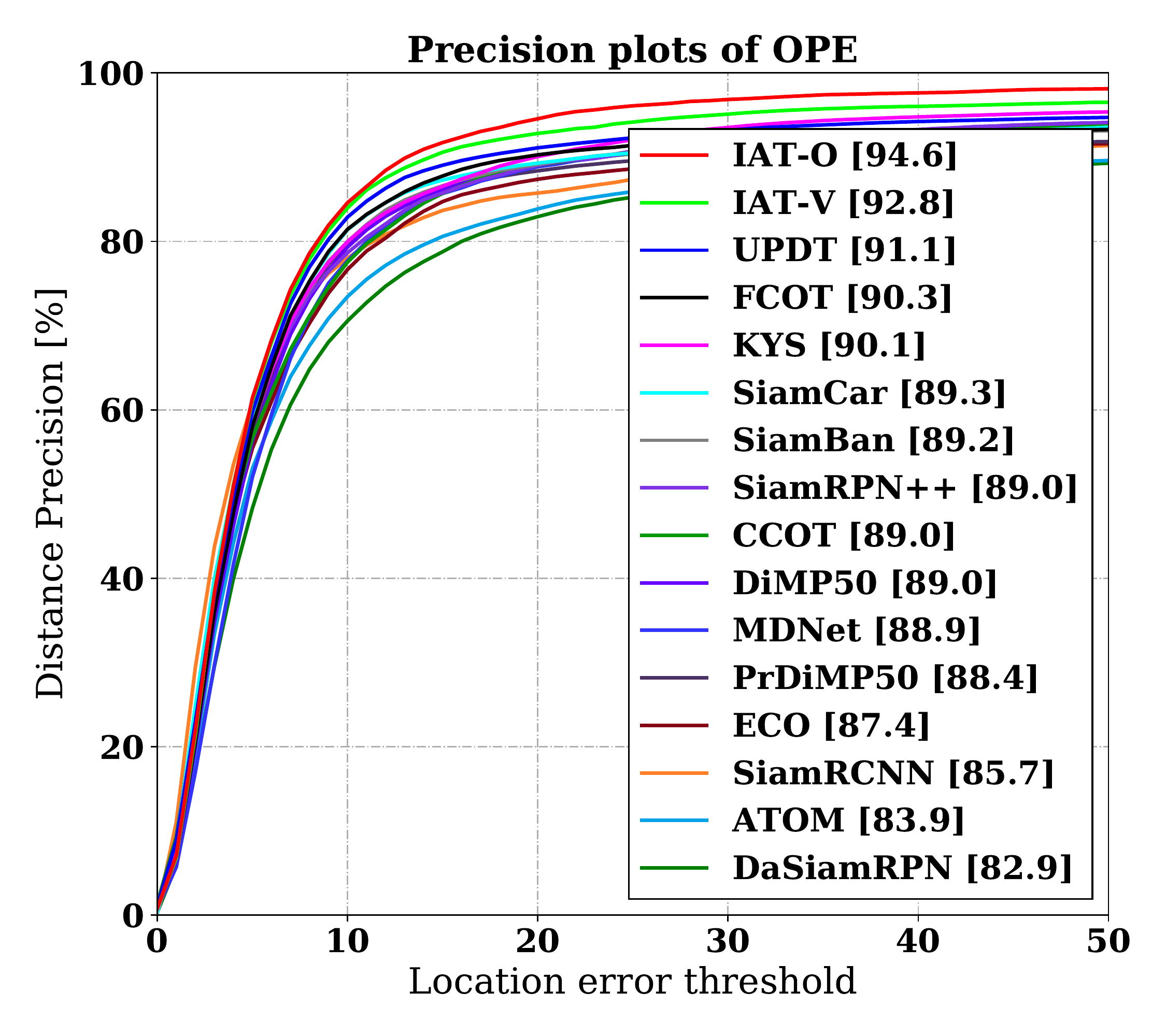}
			}
			\centerline{
				\includegraphics[width=1.0\linewidth]{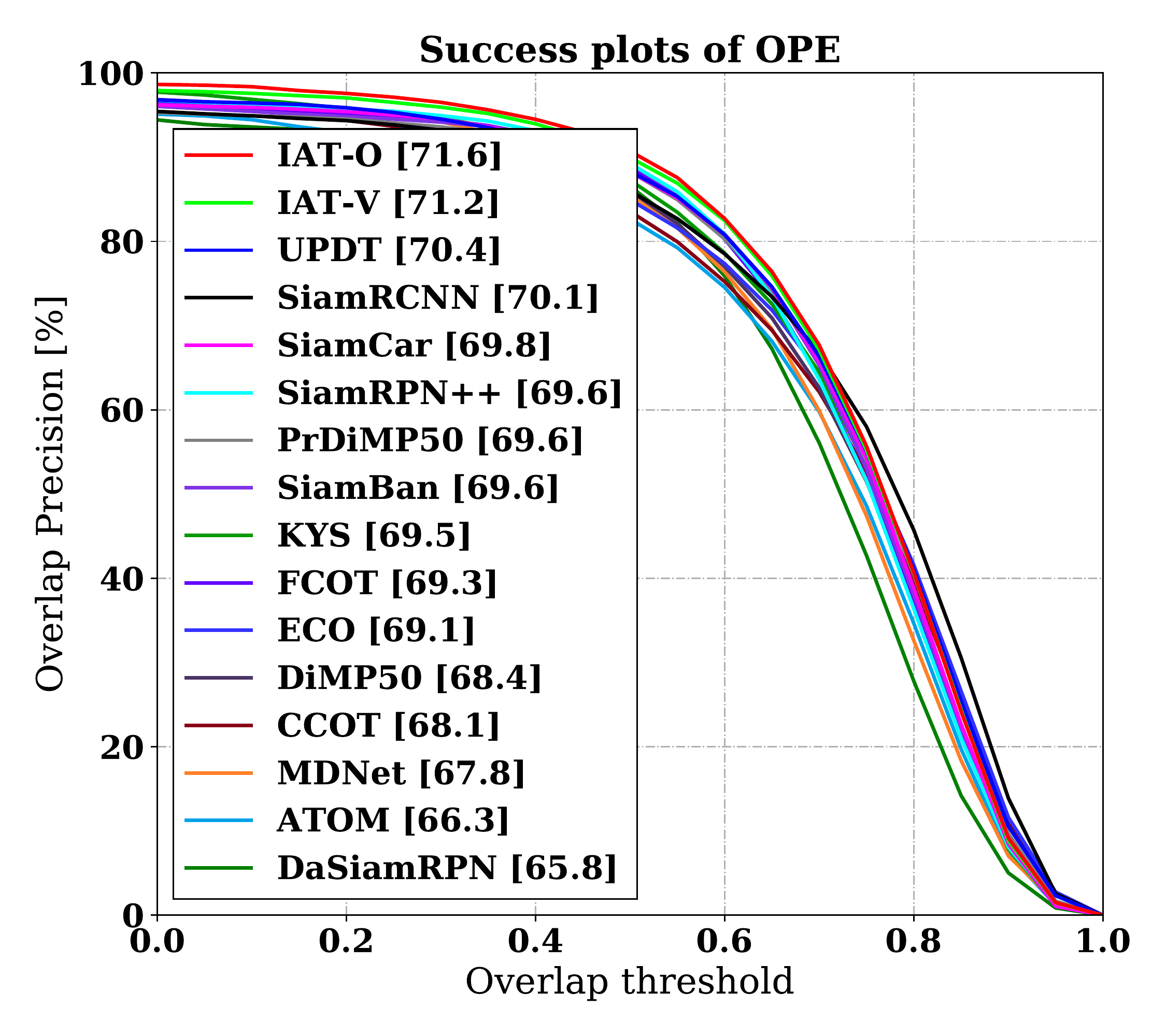}
				\includegraphics[width=1.0\linewidth]{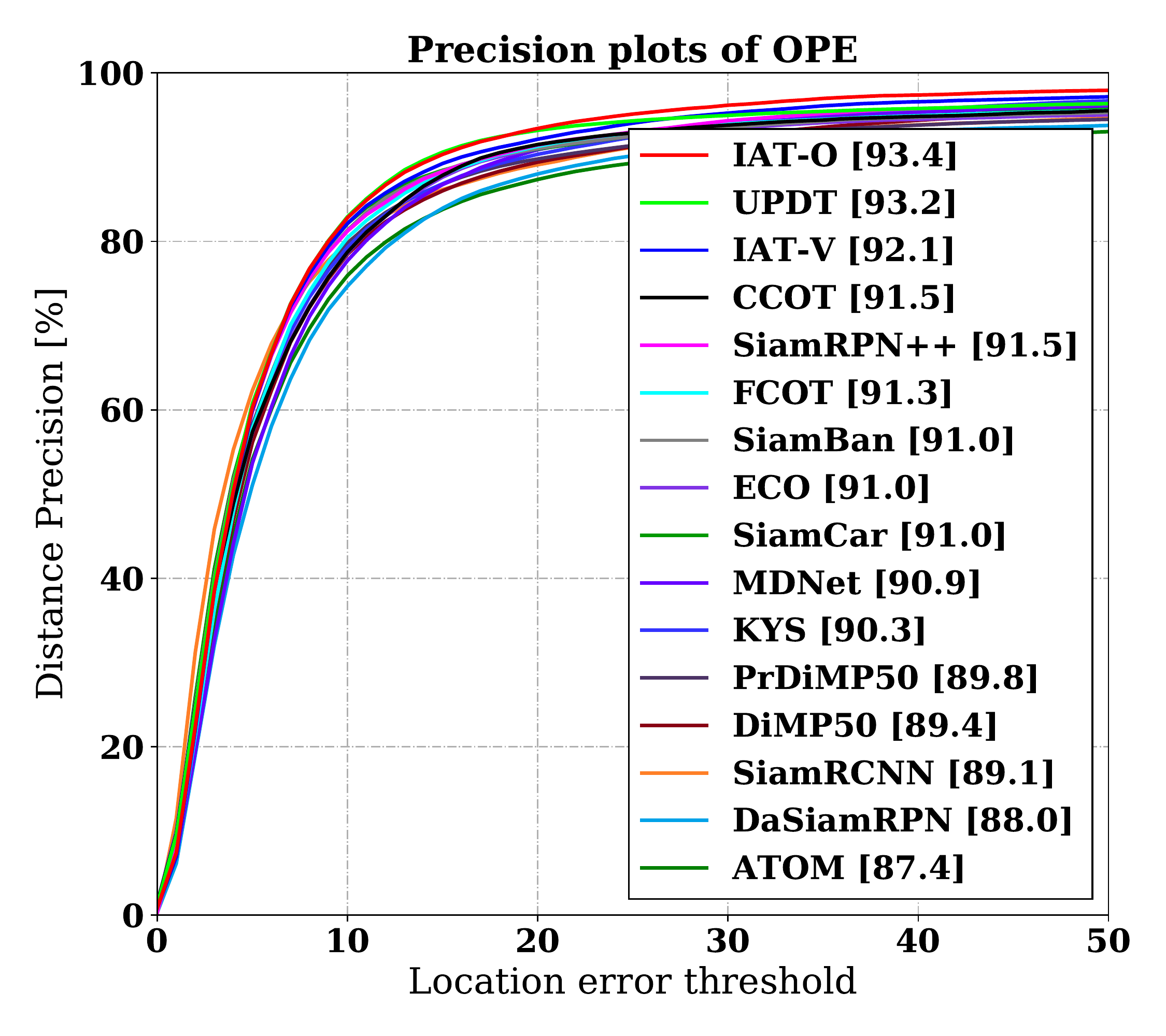}
			}
		\end{minipage}
		\caption{State-of-the-art comparison on the \textbf{OTB50} (above) and \textbf{OTB100} (below). Best view in color with zooming in}
		\label{fig:otb}
	\end{figure}
	\begin{figure*}
		\centering
		\begin{minipage}{0.38\linewidth}
			\centering
			\centerline{
				\includegraphics[width=1.0\linewidth]{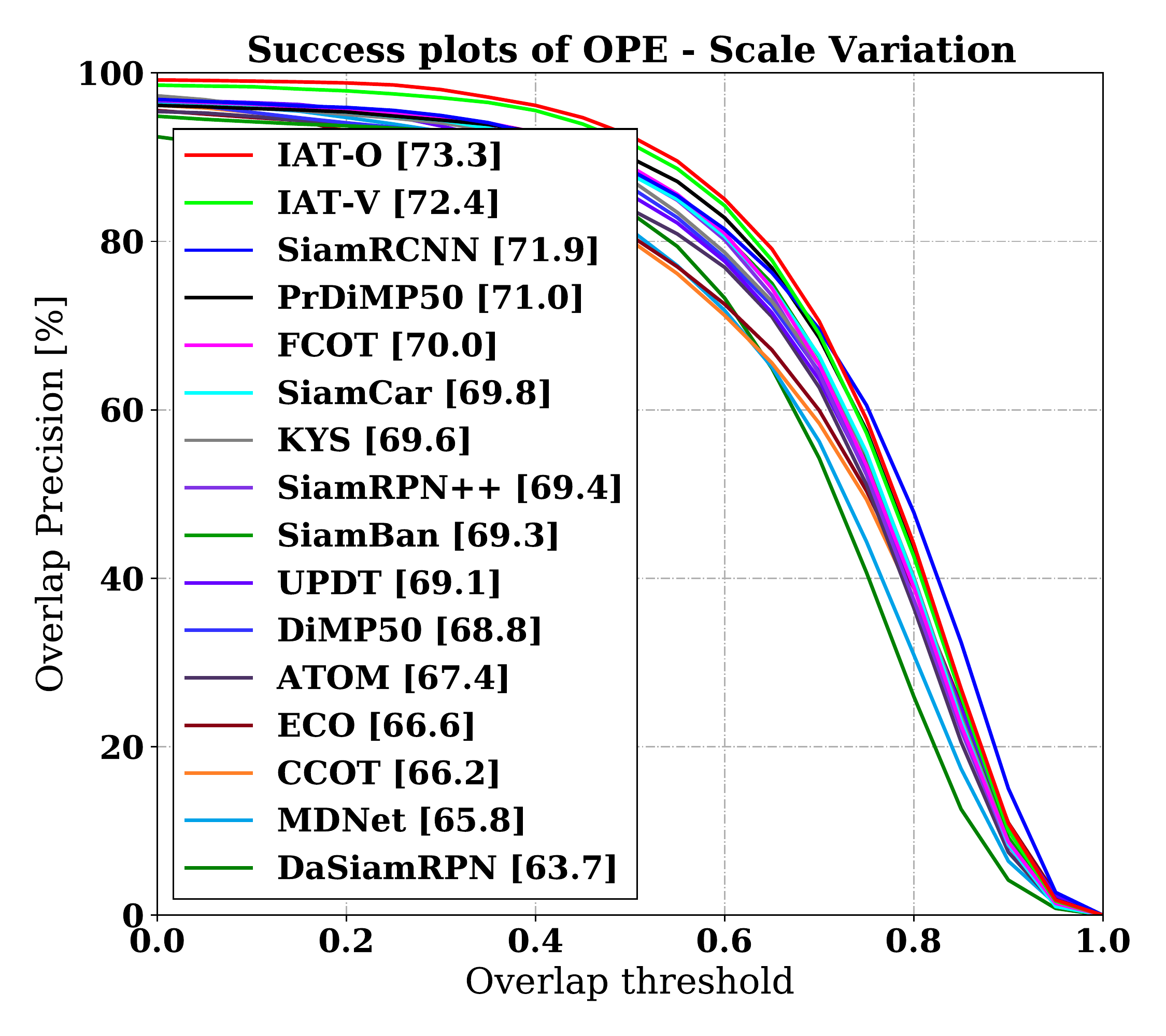}
				\includegraphics[width=1.0\linewidth]{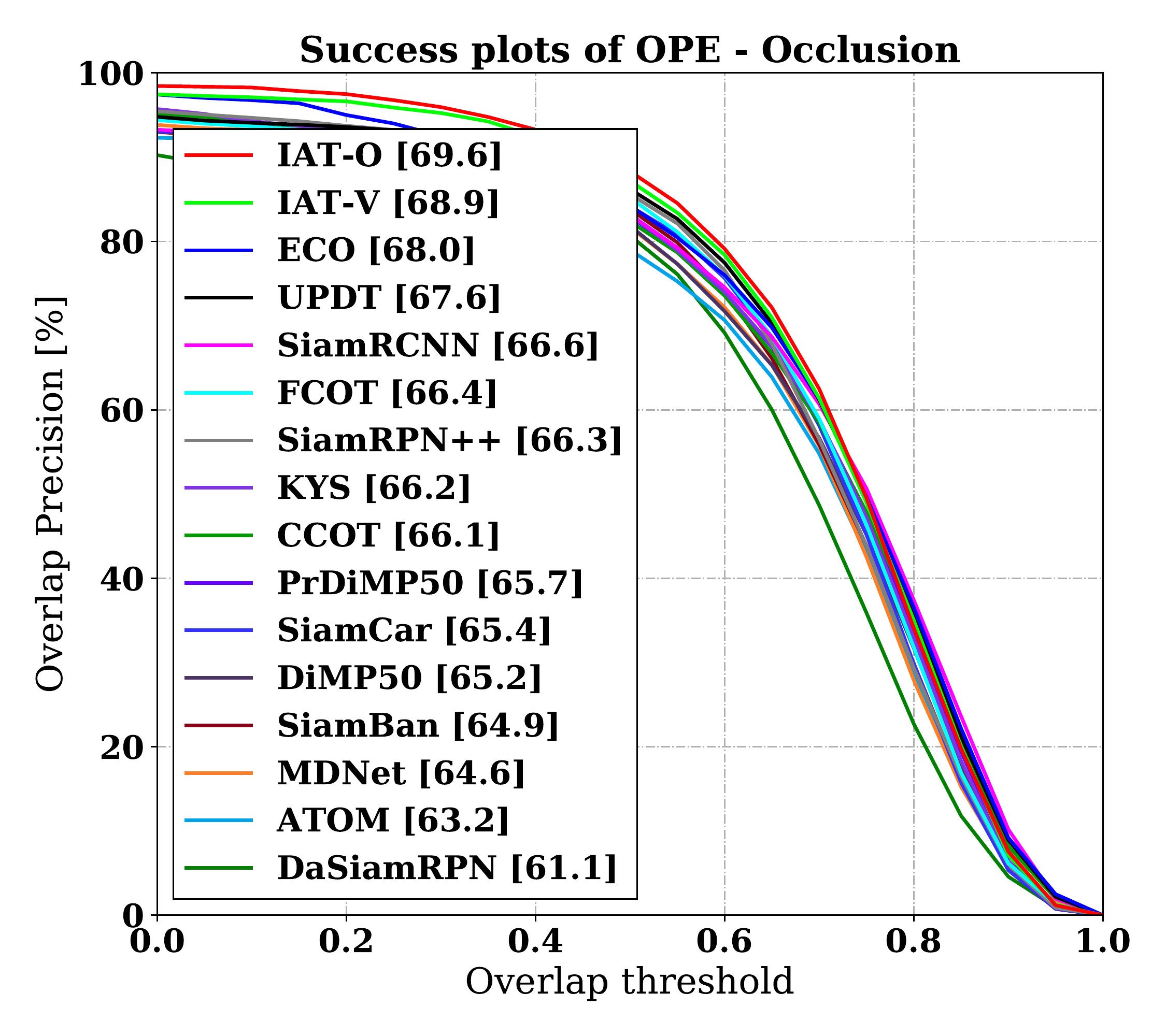}}
		    \centerline{
				\includegraphics[width=1.0\linewidth]{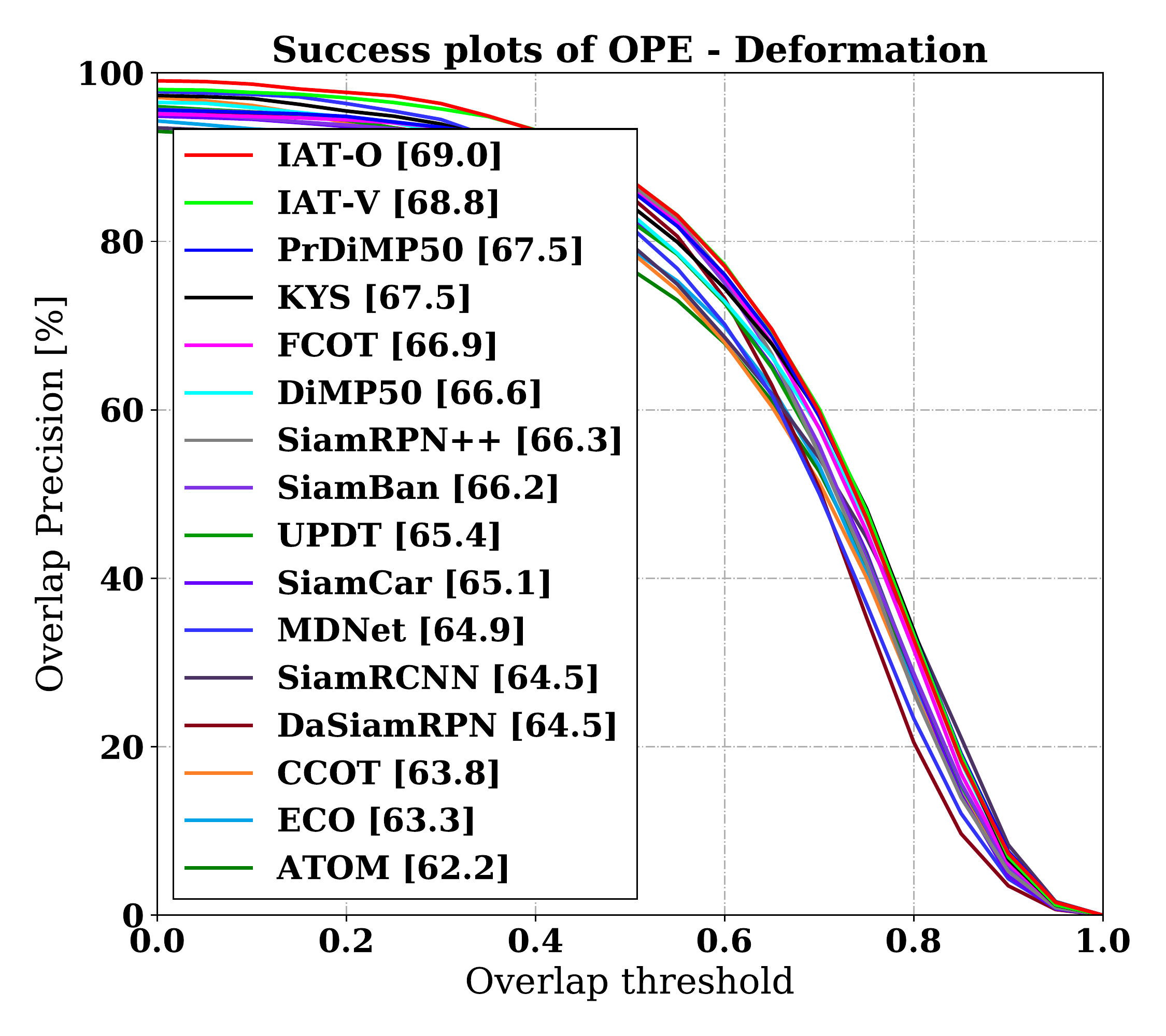}
				\includegraphics[width=1.0\linewidth]{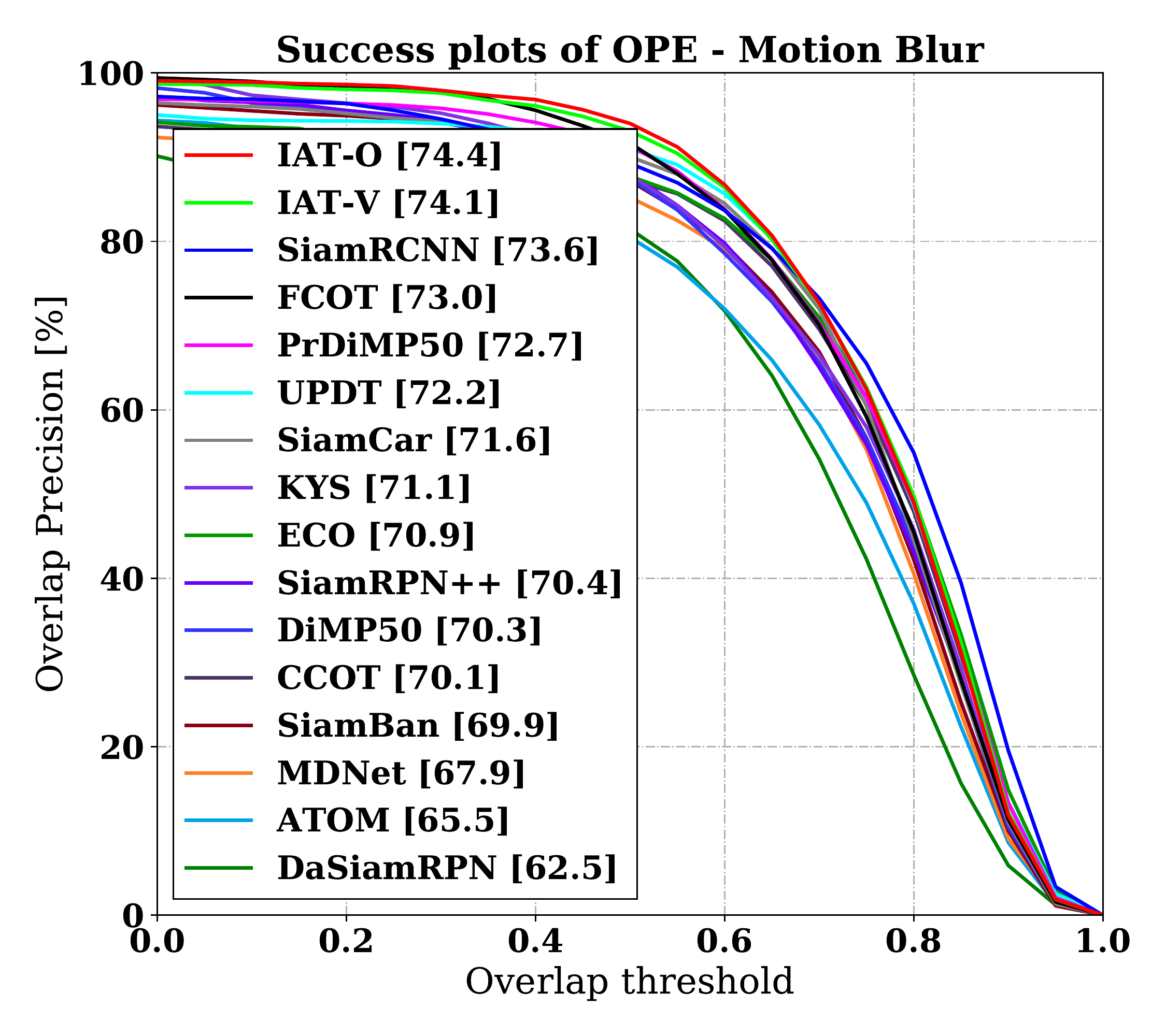}
			}
			\centerline{
				\includegraphics[width=1.0\linewidth]{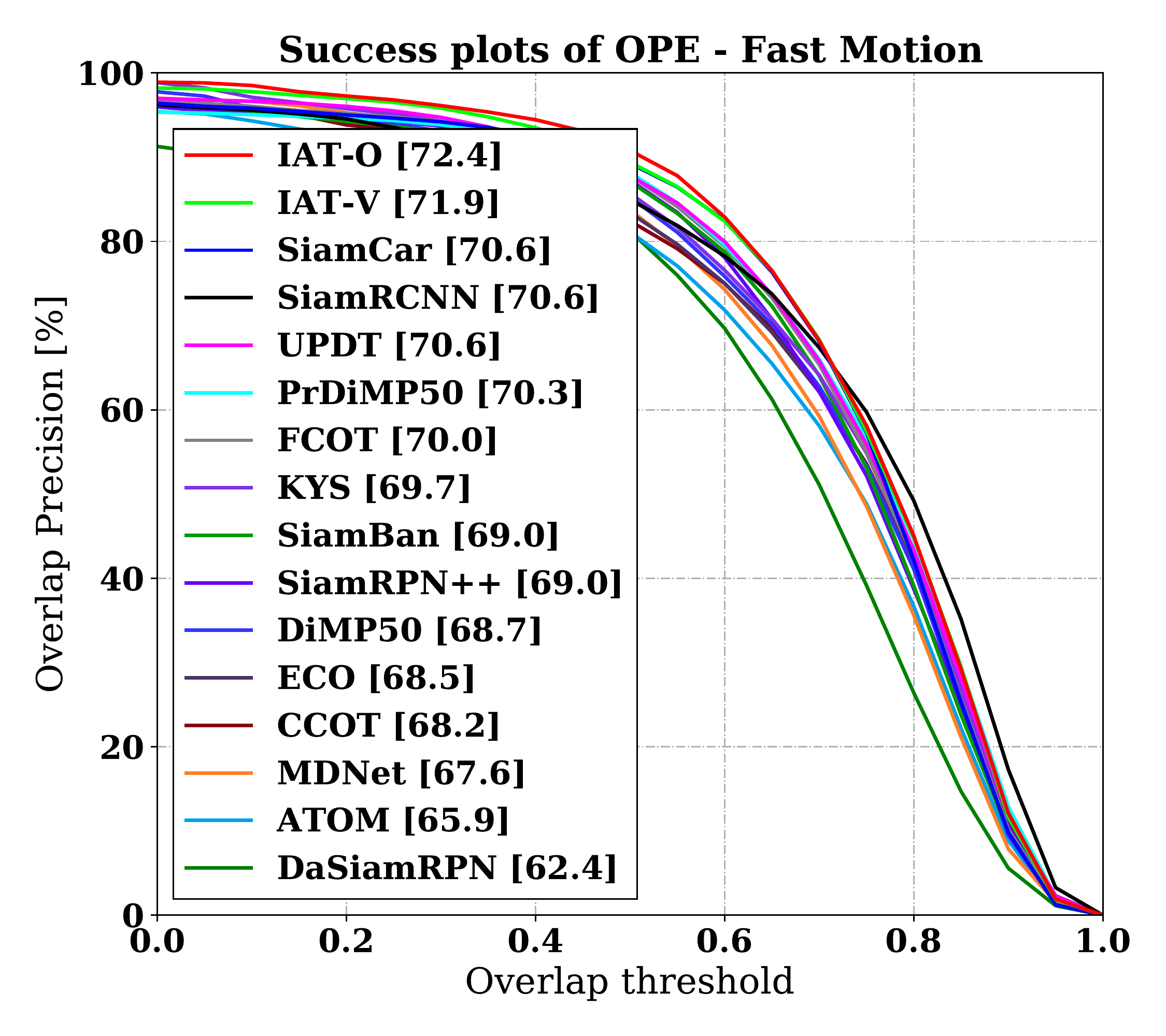}
				\includegraphics[width=1.0\linewidth]{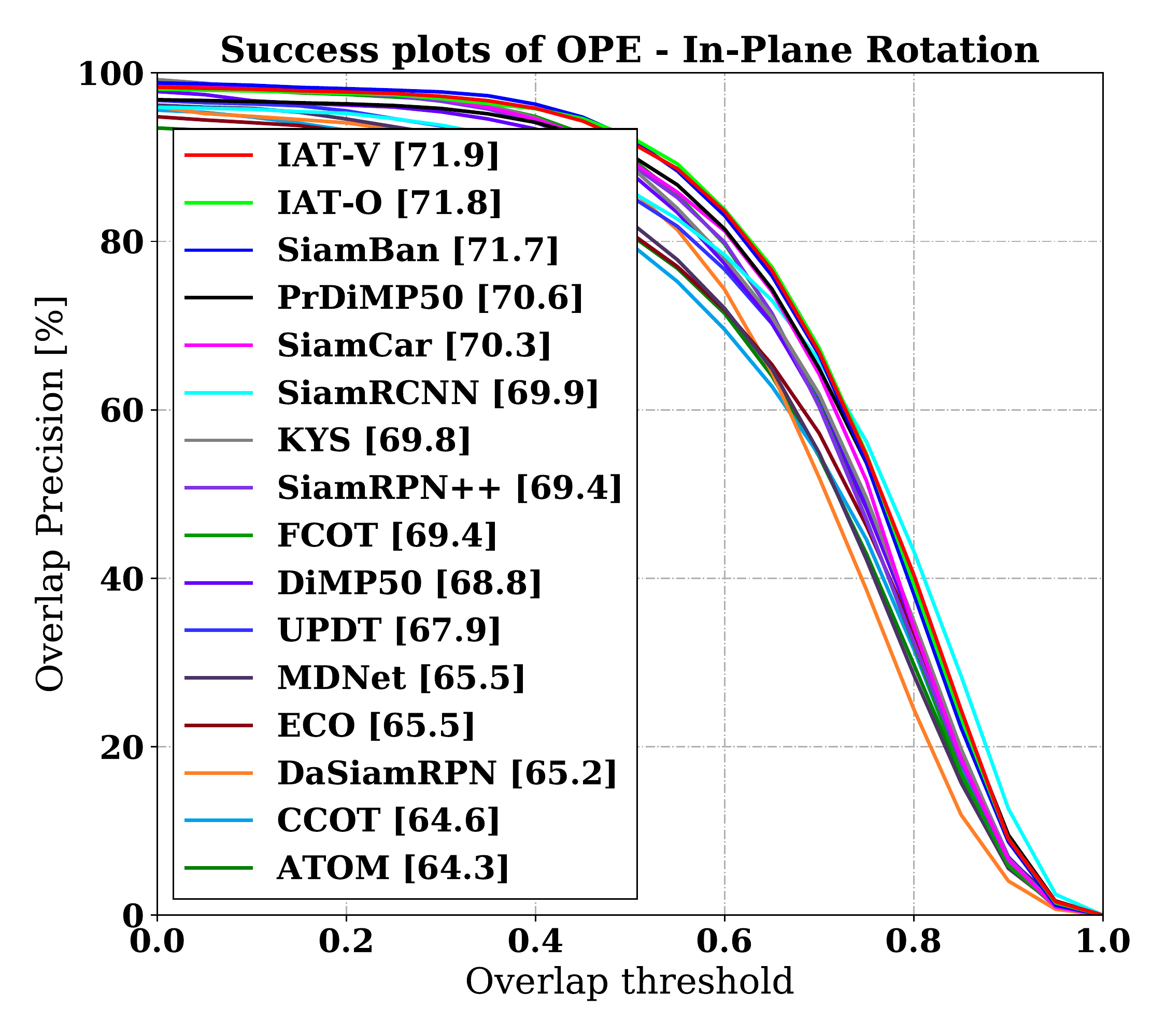}}
			\centerline{
				\includegraphics[width=1.0\linewidth]{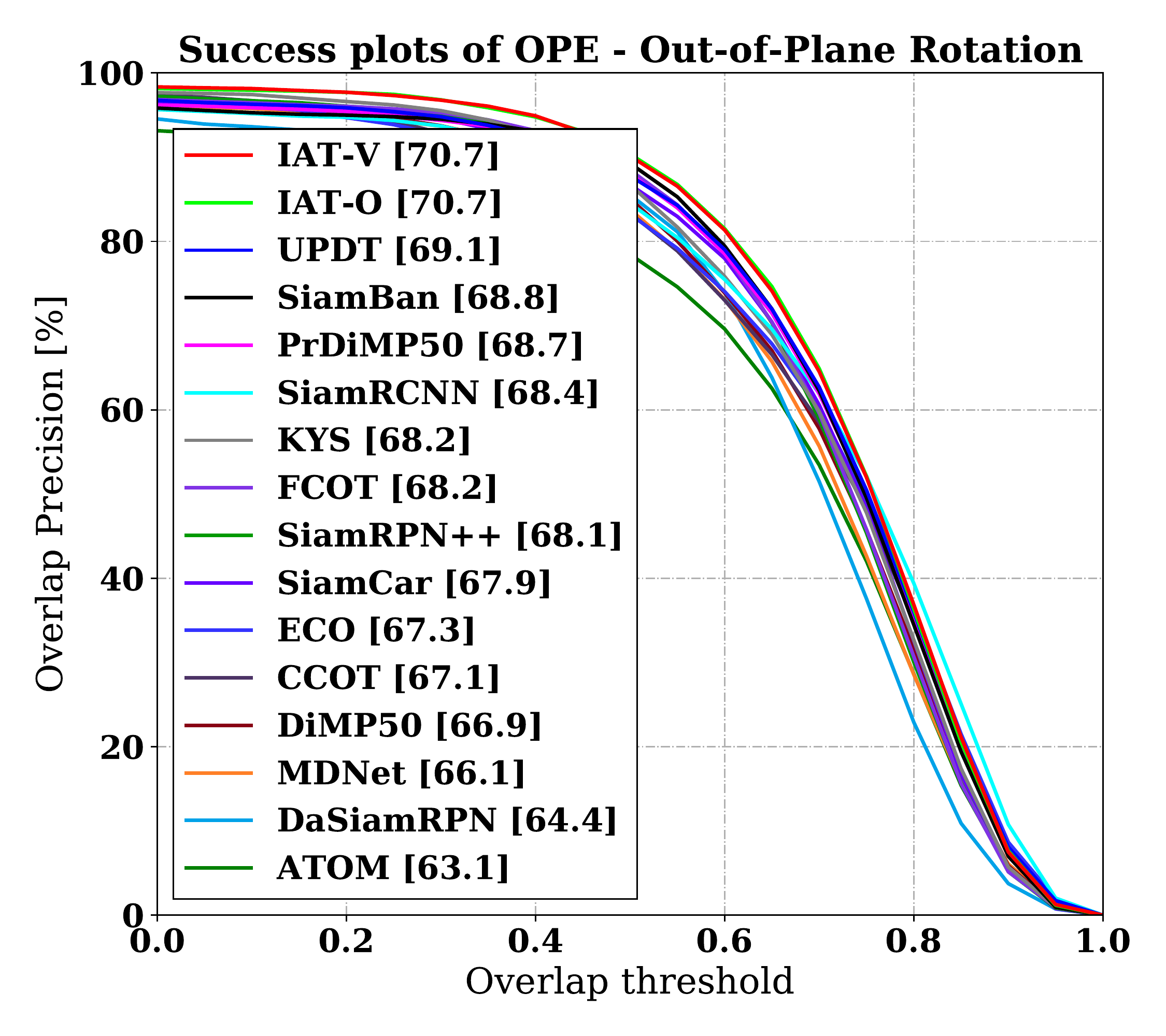}
				\includegraphics[width=1.0\linewidth]{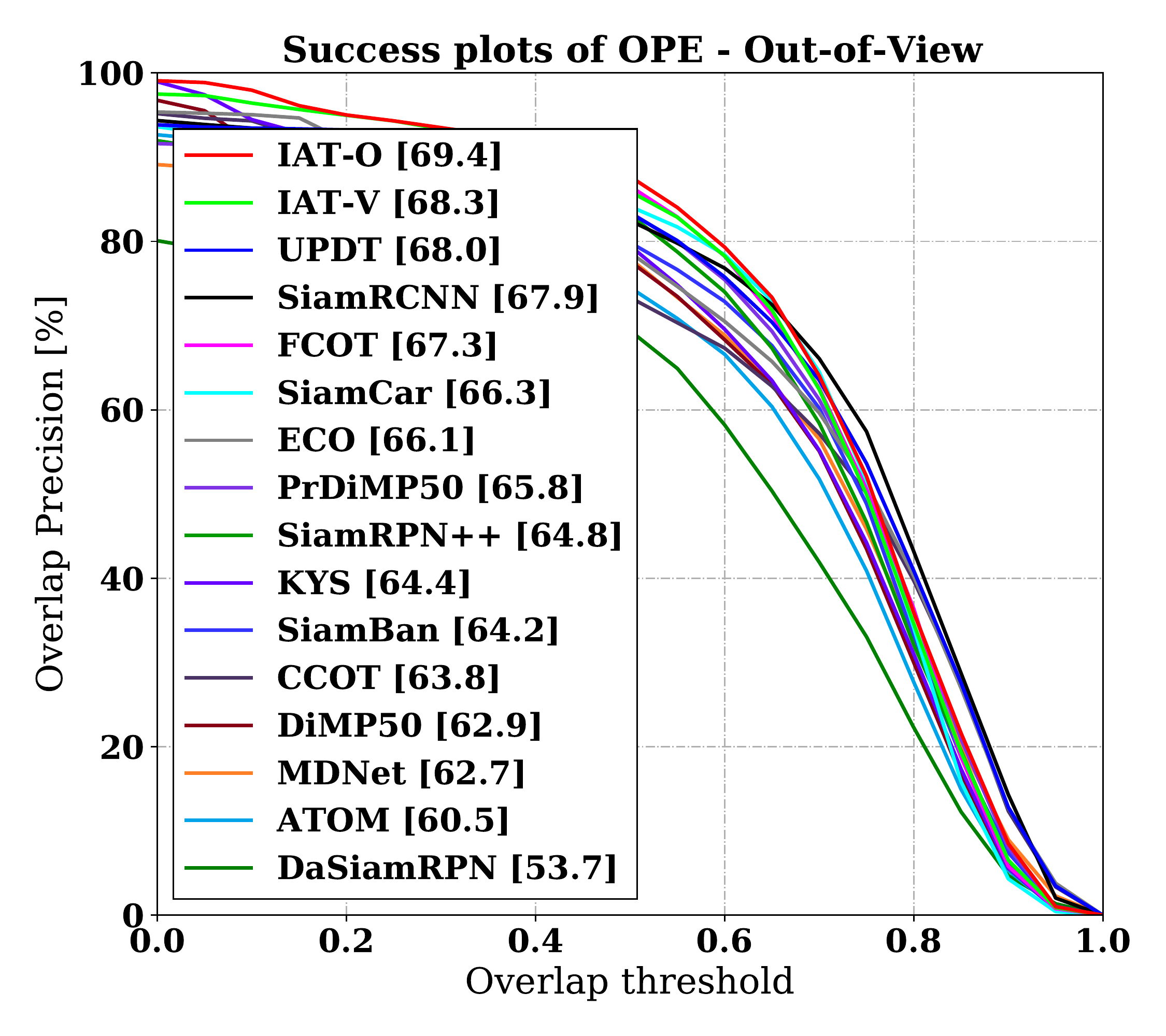}
			}
		\end{minipage}
		\caption{Evaluation on OTB100 with various challenging aspects. Our IAT-V and IAT-O achieve the best success results against the impacts of all these attributions. Best view in color with zooming in.}
		\label{fig:otb_a}
	\end{figure*}

	\subsection{Comparison with State-of-the-art Trackers}\label{sota}
	We evaluate our IAT-V and IAT-O on 8 benchmarks and compare our results with the state-of-the-art (SOTA) trackers. Both versions of our IAT obtain leading results and improve the baseline PrDiMP by a large margin on all the datasets. Especially, IAT-O achieves better results on most datasets than IAT-V, which is in line with our expectations. The reason is that IAT-O models the discriminability more fine-grained and further improves the ability to conquer the distractor. \\
	
	\subsubsection{Evaluation on OTB50 dataset}
	OTB50 contains 50 challenging videos with 11 annotated video attributes. It uses one pass evaluation (OPE) to evaluate trackers with two metrics, precision and area under curve (AUC) of the success plot. The first row in Figure~\ref{fig:otb} presents the quantitative results on this dataset. Both our IAT-V and IAT-O outperform all the other trackers. For success plot, compared to baseline PrDiMP~\cite{danelljan2020probabilistic}, our IAT-V and IAT-O improve 2.5\% and 3.6\% by a large margin respectively, and surpass the third ranking tracker UPDT~\cite{bhat2018unveiling} with 2.2\% and 3.3\%. In term of precision plot, our IAT-V and IAT-O surpass the third tracker UPDT~\cite{bhat2018unveiling} with 1.7\% and 3.5\%, and dramatically outperform PrDiMP with 4.4\% and 6.2\%. \\
	
	\subsubsection{Evaluation on OTB100 dataset}
\begin{figure}[H]
		\begin{center}
			\includegraphics[width=1\linewidth]{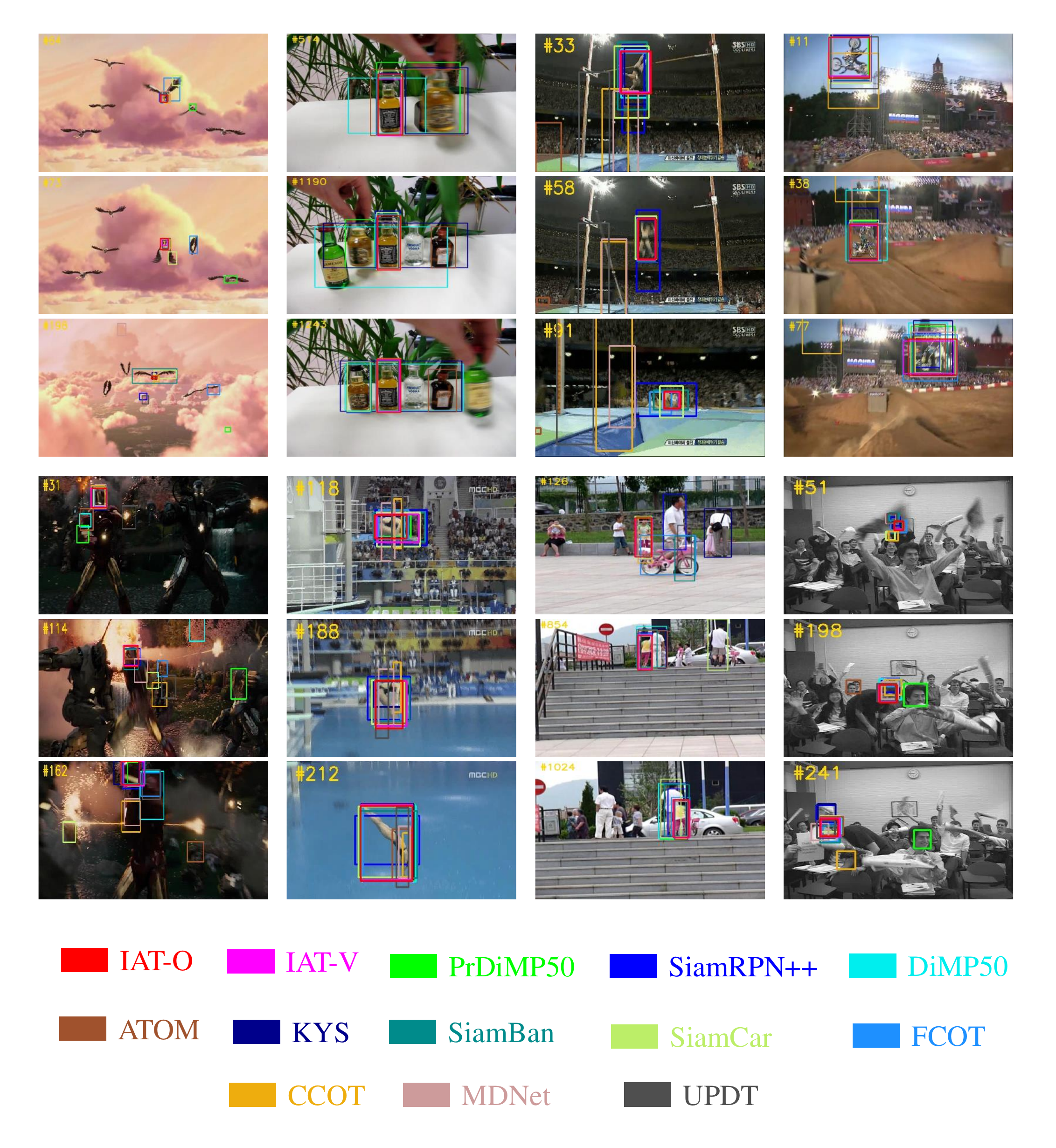}
		\end{center}
		\caption{The qualitative results of the proposed IAT-O, IAT-V on 8 challenging sequences in OTB100. Our trackers perform superior than other representative methods.}
		\label{fig:box1}
	\end{figure}

	OTB100 contains other 50 challenging videos and OTB50. The evaluation on this dataset is consistent with OTB50. The second row of Figure~\ref{fig:otb} shows the quantitative results on OTB100 dataset. The two variants have the leading results on this dataset, concretely, surpassing the following UPDT~\cite{bhat2018unveiling} by 0.8\% and 1.2\%, and exceeding the baseline model PrDiMP by 0.6\% and 2.0\% on the success plot. For precision plot, our IAT-V and IAT-O gain a great improvement of 2.3\% and 3.6\% compared to the PrDiMP. In Figure~\ref{fig:otb_a}, we analyze the proposed IAT under 8 different attributes on OTB100. The challenging aspects including scale variation, occlusion, deformation, motion blur, fast motion, in-plane rotation, out-of-plane rotation, out-of-view. Both IAT-V and IAT-O rank the top two on all the listed attributes, which demonstrates that our approaches have powerful feature representations to handle various challenges. Besides, our two variants outperform the PrDiMP on all listed attributes, which further validate the effectiveness of the proposed instance-aware strategy. Specifically, we present a qualitative evaluation on this dataset as shown in Figure \ref{fig:box1}. We compare our two categories of instance-aware trackers with several related methods: PrDiMP50, DiMP50, ATOM, KYS, SiamRPN++, SiamBan, SiamCar, FCOT, CCOT, UPDT and MDNet on eight video sequences. The results demonstrate that our trackers perform well on many challenging scenes like distractors, occlusions, illumination variations, deformations, fast motion, background clutters and in-plane rotation with our instance-aware tracking framework.\\
	
	\begin{table}[h]
		\begin{center}
			\begin{tabular}{c|c c c c }
				\hline
				Method & AO & $SR_{0.5}$ & $SR_{0.75}$ \\
				\hline
				SiamFC~\cite{bertinetto2016fully} & 0.374 & 0.328 & 0.144 \\
				SiamFC++~\cite{xu2020siamfc++} & 0.595 & 0.695 & 0.479 \\
				MDNet~\cite{nam2016learning} & 0.299 & 0.303 & 0.099 \\ 
				ECO~\cite{danelljan2017eco} & 0.316 & 0.099 & 0.111 \\
				CCOT~\cite{danelljan2016beyond} & 0.325 & 0.309 & 0.107 \\ 
				SiamRPN++~\cite{li2019siamrpn++} & 0.517 & 0.593 & 0.325 \\
				SiamCar~\cite{guo2020siamcar} & 0.569 & 0.670 & 0.415 \\
				D3S~\cite{lukezic2020d3s} & 0.597 & 0.676 & 0.462 \\
				KYS~\cite{bhat2020know} & 0.636 & 0.751 & 0.515 \\
				DiMP50~\cite{bhat2019learning} & 0.611 & 0.717 & 0.492 \\
				PrDiMP50~\cite{danelljan2020probabilistic} & 0.634 & 0.738 & 0.543 \\
				FCOT~\cite{cui2020fully} & 0.634 & 0.766 & 0.521 \\
				Ocean~\cite{zhang2020ocean} & 0.611 & 0.721 & - \\
				ROAM++~\cite{yang2020roam} & 0.465 & 0.532 & 0.236 \\ 
				SiamRCNN~\cite{voigtlaender2020siam} & 0.649 & - & -\\
				\hline
				\textbf{IAT-V} & \textbf{0.676} & \textbf{0.792} & \textbf{0.583}\\
				\textbf{IAT-O} & \textbf{0.685} & \textbf{0.808} & \textbf{0.598} \\ 
				\hline
			\end{tabular}
		\end{center}
		\caption{State-of-the-art comparison on \textbf{GOT-10k} test set in terms of average overlap(AO), and success rates(SR) at overlap thresholds 0.5 and 0.75. The best two results are highlighted by bold.}
		\label{tab:GOT-10k}
	\end{table}
	\begin{table}[t]
		\begin{center}
			\scalebox{1.0}{\begin{tabular}{c|c c c c }
				\hline
				Method & Success & Prec. & Norm Prec. \\
				\hline
				SiamFC++~\cite{xu2020siamfc++} & 0.544 & - & - \\
				DaSiamRPN~\cite{zhu2018distractor} & 0.415 & - & 0.496 \\ 
				MDNet~\cite{nam2016learning} & 0.397 & - & 0.460 \\ 
				UpdateNet~\cite{zhang2019learning} & 0.495 & - & 0.569 \\ 
				SiamRPN++~\cite{li2019siamrpn++} & 0.496 & 0.491 & 0.569  \\
				ATOM~\cite{danelljan2019atom} & 0.515 & - & 0.576  \\
				DiMP50~\cite{bhat2019learning} & 0.569 & - & 0.643  \\
				PrDiMP50~\cite{danelljan2020probabilistic} & 0.598 & - & 0.688 \\
				SiamAtten~\cite{yu2020deformable} & 0.560 & - & 0.648 \\
				SiamCar~\cite{guo2020siamcar} & 0.507 & 0.510 & 0.600 \\
				FCOS-MAML~\cite{wang2020tracking} & 0.523 & - & -  \\
				FCOT~\cite{cui2020fully} & 0.569 & - & 0.678 \\
				CGACD~\cite{du2020correlation} & 0.518 & - & 0.626 \\
				Ocean~\cite{zhang2020ocean} & 0.560 & - & 0.566 \\
				ROAM++~\cite{yang2020roam} & 0.447 & 0.445 & - \\
				\hline
				\textbf{IAT-V} & \textbf{0.627} & \textbf{0.648} & \textbf{0.715}\\
				\textbf{IAT-O} & \textbf{0.634} & \textbf{0.654} & \textbf{0.720} \\
				\hline
			\end{tabular}}
		\end{center}
		\caption{State-of-the-art comparison on \textbf{LaSOT} dataset in terms of success, precision, and normalized precision. The best two results are highlighted by bold.}
		\label{tab:lasot}
	\end{table}

	\subsubsection{Evaluation on GOT-10k dataset}
	GOT-10k is a large highly-diverse dataset that contains over 10000 video segments and has 180 test videos. Its train and test splits have no overlap in object classes, thus overfitting on particular classes is avoided. In addition, this benchmark requires all trackers to use only the train split for model training while external datasets are forbidden. We strictly follow this protocol and evaluate our approach using the online evaluation server on the test split consisting of 180 sequences. Table~\ref{tab:GOT-10k} shows the result in terms of average overlap (AO) and success rates (SR) at overlap thresholds 0.5 and 0.75. Both our trackers surpass all other trackers in all metrics. In particular, our IAT-V surpasses the baseline PrDiMP~\cite{danelljan2020probabilistic} with an improvement of 4.2\%, 5.4\% and 4.0\% on three metrics, respectively. Our IAT-O exceed the PrDiMP with 5.1\%, 7.0\% and 5.5\% on three metrics, respectively. Furthermore, IAT-O outperforms the third best tracker SiamRCNN~\cite{voigtlaender2020siam} with a gap of 3.6\% on AO. \\

	\begin{table}[t]
		\begin{center}
			\scalebox{1}{\begin{tabular}{c|c c c c }
				\hline
				Method & Success & Prec. & Norm Prec. \\
				\hline
				SiamFC~\cite{bertinetto2016fully} & 0.571 &  0.533 & 0.663 \\ 
				SiamFC++~\cite{xu2020siamfc++} & 0.754 & 0.705 & 0.800 \\
				SiamAtten~\cite{yu2020deformable} & 0.752 & - & 0.817\\
				SiamMask~\cite{wang2019fast} & 0.725 & 0.664 & 0.778 \\ 
				DaSiamRPN~\cite{zhu2018distractor} & 0.638 & 0.591 & 0.733 \\ 
				SiamRPN++~\cite{li2019siamrpn++} & 0.733 & 0.694 & 0.800 \\
				MDNet~\cite{nam2016learning} & 0.606 & 0.565 & 0.705 \\ 
				UPDT~\cite{bhat2018unveiling} & 0.611 & 0.557 & 0.702  \\ 
				
				ATOM~\cite{danelljan2019atom} & 0.703 & 0.648 & 0.771 \\
				DiMP50~\cite{bhat2019learning} & 0.740 & 0.687 & 0.801 \\
				PrDiMP50~\cite{danelljan2020probabilistic} & 0.758 & 0.704 & 0.816 \\
				D3S~\cite{lukezic2020d3s} & 0.728 & 0.664 & 0.768 \\
				KYS~\cite{bhat2020know} & 0.740 & 0.688 & 0.800 \\
				FCOS-MAML~\cite{wang2020tracking} & 0.757 & - & 0.822 \\
				CGACD~\cite{du2020correlation} & 0.711 & 0.693 & 0.800 \\
				ROAM++~\cite{yang2020roam} & 0.670 & 0.623 & 0.754 \\
				AFFT-Res50~\cite{zhao2021adaptive} &0.732 & 0.680 &0.794 \\
				\hline
				\textbf{IAT-V} & \textbf{0.780} & \textbf{0.728} & \textbf{0.829}\\
				\textbf{IAT-O} & \textbf{0.777} & \textbf{0.726} & \textbf{0.827}\\
				\hline
			\end{tabular}}
		\end{center}
		\caption{State-of-the-art comparison on the \textbf{TrackingNet} test set in terms of success, precision, and normalized precision. The best two results are highlighted by bold.}
		\label{tab:trackingnet}
	\end{table}  
	\begin{table}[h]
		\begin{center}
			\scalebox{1}{
				\begin{tabular}{c|c c c c c c c}
					\hline
					Method & EAO & Accuracy & Robustness \\
					\hline
					SiamRPN++~\cite{li2019siamrpn++} & 0.292 & 0.580 & 0.446 \\
					SiamMask~\cite{wang2019fast} & 0.287 & 0.594 & 0.461 \\ 
					ATOM~\cite{danelljan2019atom} & 0.301 & 0.603 & 0.411 \\
					PrDiMP50~\cite{danelljan2020probabilistic} & 0.304 & 0.616 & 0.398 \\
					SiamBan~\cite{chen2020siamese} & \textbf{0.327} & 0.602 & 0.396 \\
					Retina-MAML~\cite{wang2020tracking} & 0.313 & 0.570 & \textbf{0.366} \\
					Ocean~\cite{zhang2020ocean} & \textbf{0.350} & 0.594 & \textbf{0.316} \\
					\hline
					\textbf{IAT-V} & 0.312 & \textbf{0.629} & 0.391 \\
					\textbf{IAT-O} & 0.320 & \textbf{0.635} & 0.386 \\
					\hline
			\end{tabular} }
		\end{center}
		\caption{State-of-the-art comparison on \textbf{VOT2019} dataset in terms of expected average overlap(EAO), accuracy and robustness(tracking failure). The best two results are highlighted by bold.}
		\label{tab:vot2019}
	\end{table}

	\subsubsection{Evaluation on LaSOT dataset}
	This dataset provides a large-scale, high-quality dense annotations with 1,400 videos in total and 280 videos in the test set. LaSOT has 70 categories of objects with each containing twenty sequences, and the average video length is 2512 frames, which is sufficient to evaluate long-term trackers. LaSOT also adopts OPE to evaluate success and precision similar to OTB. We evaluate our approach on the test split. Table~\ref{tab:lasot} shows the result in terms of success, precision and normalized precision. Our IAT-O achieves the best results, specifically, 63.4\% on success score, 65.4\% on precision score and 72.0\% on norm precision. Our baseline PrDiMP~\cite{danelljan2020probabilistic} is the third best tracker. Both IAT-V and IAT-O outperform the PrDiMP with 2.9\% and 3.6\% on success score, 2.7\% and 3.2\% on norm precision respectively. \\

	\begin{figure*}[t]
		\centering
		\begin{minipage}[b]{0.5\linewidth}
			\centering
			\centerline{
				\includegraphics[width=1.0\linewidth]{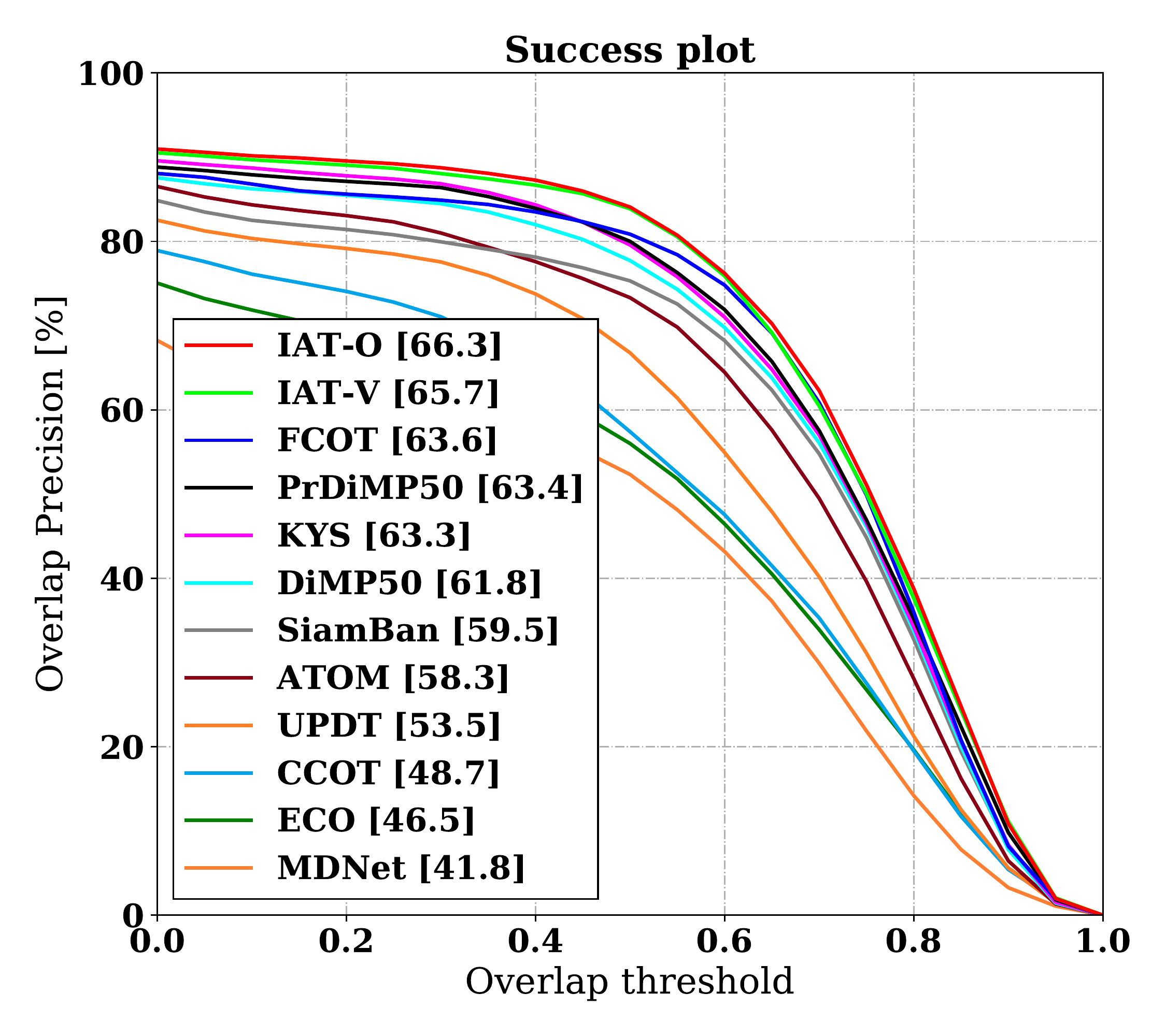}
				\includegraphics[width=1.0\linewidth]{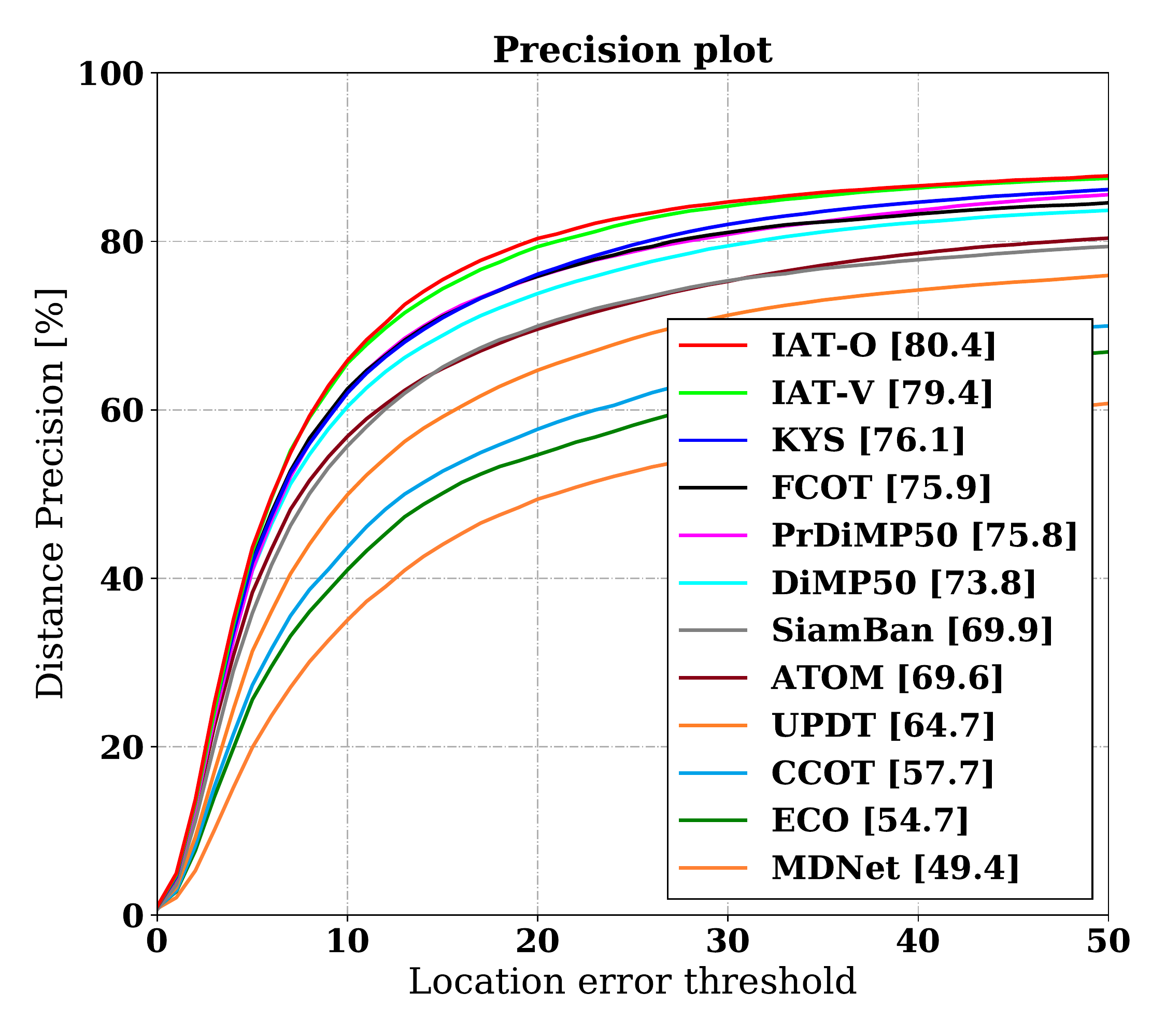}
			}
		\centerline{
				\includegraphics[width=1.0\linewidth]{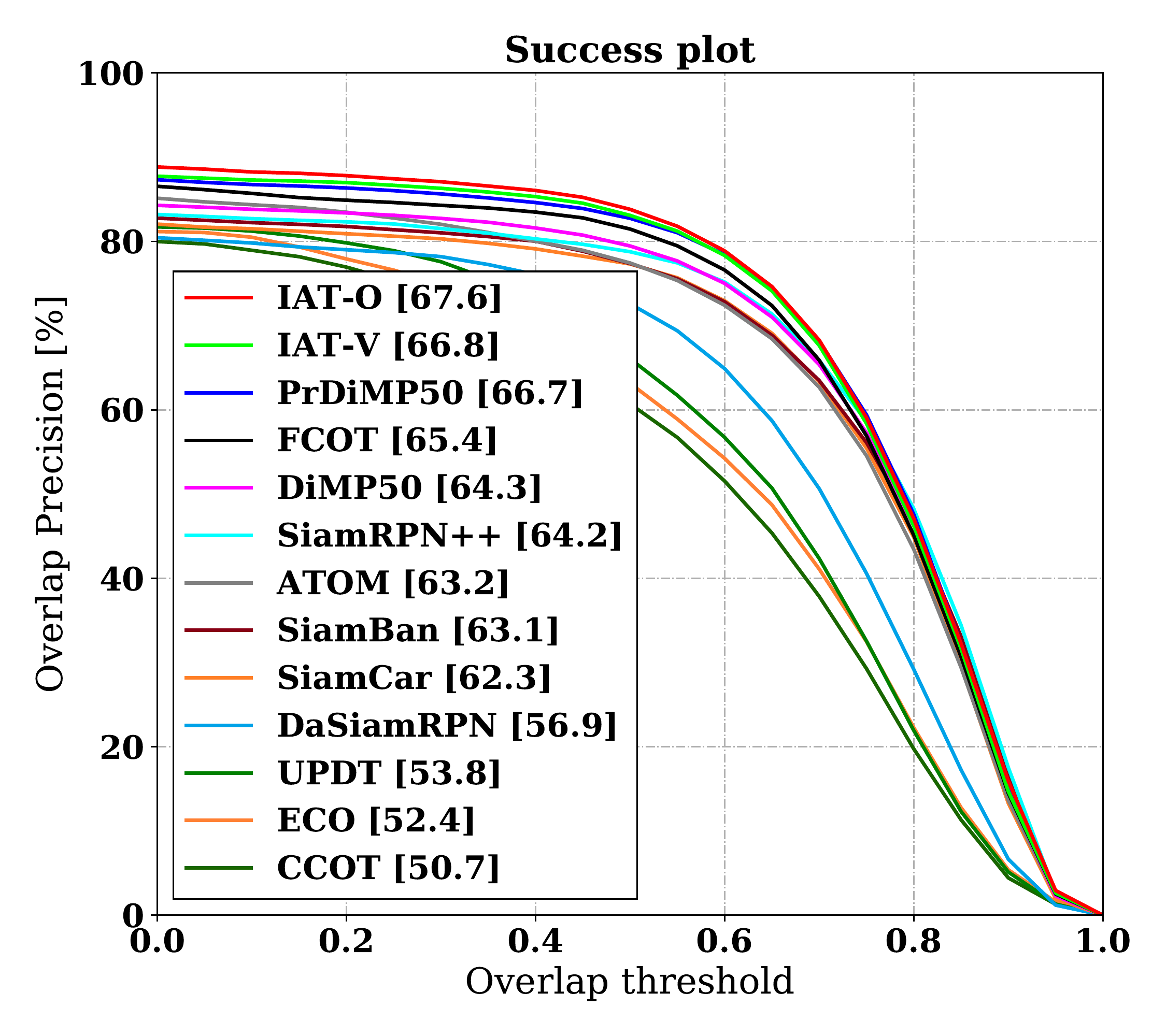}
				\includegraphics[width=1.0\linewidth]{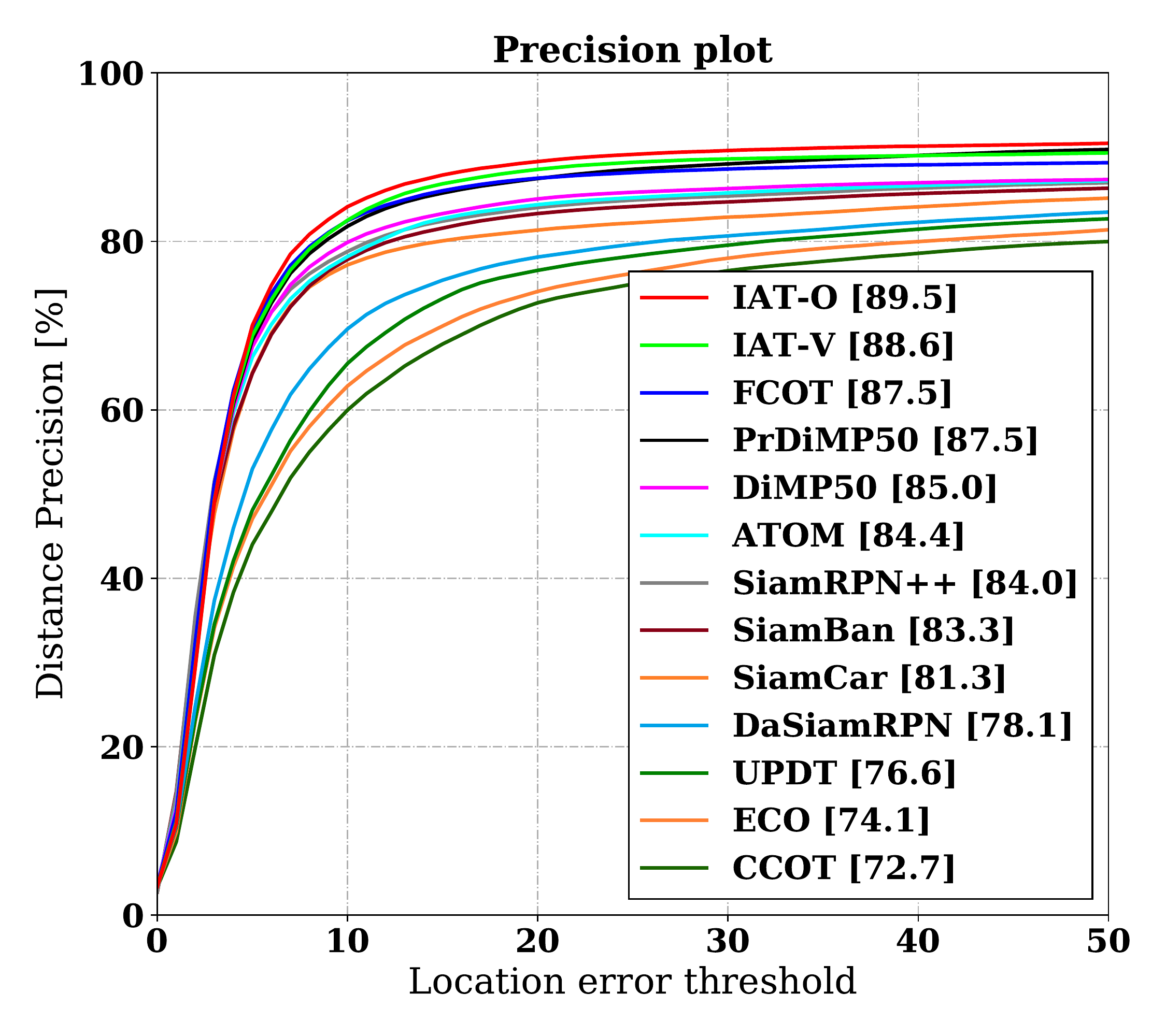}
			}
		\end{minipage}
		\caption{State-of-the-art comparison on the \textbf{NFS30} (top) and \textbf{UAV123} (bottom). Best view in color with zooming in.}
		\label{fig:nfs30}
	\end{figure*}

	\subsubsection{Evaluation on NFS dataset}
	We evaluate on the 30 FPS version of the dataset, containing challenging videos with fast-moving objects. The evaluation on this dataset is the same as OTB100. The top of Figure~\ref{fig:nfs30} shows the success plot over all the 100 videos, from which we can see that our two variants of IAT achieve a substantial improvement over the previous SOTA methods on this dataset. In terms of the AUC score of the success plot, the IAT-V reaches 65.7\% while the IAT-O realizes a better score 66.3\%, surpassing the following top-ranking tracker FCOT~\cite{cui2020fully} by 2.1\% and 2.7\%, and surpassing the baseline PrDiMP by 2.3\% and 2.9\% respectively. As for the precision score, our two variants reach 80.4\% and 79.4\% and exceed the following tracker KYS~\cite{bhat2020know} by 3.3\% and 4.3\%, outperform the PrDiMP by 3.6\% and 4.6\%, respectively. \\
	%
	
    \subsubsection{Evaluation on UAV123 dataset}
    UAV123 contains 123 sequences captured from low-altitude UAVs. Unlike other tracking datasets, the viewpoint of UAV123 is aerial and the targets to be tracked are usually small. It includes 123 videos with 915 frames average length. Its evaluation is the same as OTB100. The bottom of Figure~\ref{fig:nfs30} displays the success plot over all the 123 videos. Specifically, IAT-V and IAT-O get the top performance of 67.6\% and 66.8\% on success plot. And both IAT-V and IAT-O lead the top two results with 89.5\% and 88.6\% on precision plot, outperforming our baseline PrDiMP by 1.1\% and 2.0\%, respectively.  \\

	\subsubsection{Evaluation on TrackingNet dataset}
	This is a large-scale dataset consisting of real-world videos sampled from YouTube. It contains 30000 sequences with 14 million dense annotations and 511 sequences in the test set and covers diverse object classes and scenes, requiring trackers to have both discriminative and generative capabilities. The trackers are evaluated using an online evaluation server on the test set of 511 videos. Table~\ref{tab:trackingnet} shows the results in terms of success, precision and normalized precision. Our IAT-V and IAT-O rank the first and second among all recent SOTA trackers. IAT-V outperform the PrDiMP by 2.2\%, 2.4\% and 1.3\%, IAT-O surpass 1.9\%, 2.2\% and 1.1\%, respectively. \\
	
	\subsubsection{Evaluation on VOT2019 dataset}
	This dataset consists of 60 videos that challenge reinitialized trackers at the failure frames. It has a burn-in period of ten frames, which means ten frames after initialization will be labeled as invalid for accuracy computation. The performance is evaluated in terms of accuracy (average overlap in the course of successful tracking) and robustness (failure rate). The two measures are merged in a single metric, Expected Average Overlap (EAO), which provides the overall performance ranking. Table~\ref{tab:vot2019} shows the comparison of our approach with the SOTA trackers. IAT-V and IAT-O outperform other methods on accuracy and have comparable results on EAO and Robustness. Both IAT-V and IAT-O surpass the PrDiMP with 0.8\% and 1.6\% on EAO, 1.3\% and 1.9\% on accuracy, 0.7\% and 1.2\% on robustness respectively. \\

\section{Conclusion and Future Work}
	In this paper, we propose a novel instance-aware network for visual tracking, which explicitly excavates the discriminability of feature representations. Specifically, we propose to model the classical tracking pipeline as an instance-level classification problem. Moreover, we implement two variants of the proposed IAT tracker, i.e., a video-level one and an object-level one. The former enhances the ability to recognize the target from the background while the latter boosts the discriminative power for relieving the target-distractor dilemma. Both variants achieve leading performance on 8 challenging datasets while running at 30FPS.
	
	 We believe the architectures and ideas discussed in this paper are successful in explicit discriminability modeling. It could be extended for other related tasks such as video object segmentation. We leave this as our future work.

\section{Acknowledgments}
The authors declare that their work has not received any funding.


\bibliographystyle{plain}
\bibliography{ref.bib}

\end{document}